%% file: main.tex
\documentclass[final]{cvpr}

\usepackage{times}
\usepackage{epsfig}
\usepackage{graphicx}
\usepackage{amsmath}
\usepackage{amssymb}
\usepackage{subcaption}

\usepackage{enumitem}
\usepackage{booktabs}

\usepackage{pgfplots} %
\usepackage{pgfplotstable} %
\pgfplotsset{compat=newest} %
\usetikzlibrary{plotmarks} %
\usetikzlibrary{colorbrewer} %

\usepackage{xspace}

\usepackage{algorithm}
\usepackage{algpseudocode}
\usepackage{multirow}
\usepackage{tabularx}
\usepackage{wrapfig}
\usepackage{stfloats}  %

\makeatletter
\DeclareRobustCommand\onedot{\futurelet\@let@token\@onedot}
\def\@onedot{\ifx\@let@token.\else.\null\fi\xspace}

\usepackage[pagebackref=true,breaklinks=true,colorlinks,bookmarks=false]{hyperref}

\def\eg{\emph{e.g}\onedot} 
\def\ie{\emph{i.e}\onedot}

\usepackage{pifont}

\def\cvprPaperID{0000} %
\def\confYear{CVPR 0000}

\begin{document}

\title{DeepLab2: A TensorFlow Library for Deep Labeling}

\author{
Mark Weber\textsuperscript{1}\thanks{The first three authors contributed equally to this work.}~~~
Huiyu Wang\textsuperscript{2}\footnotemark[1]~~~
Siyuan Qiao\textsuperscript{2}\footnotemark[1]~~~
Jun Xie\textsuperscript{4}~~~
Maxwell D. Collins\textsuperscript{4}\\
Yukun Zhu\textsuperscript{4}~~~
Liangzhe Yuan\textsuperscript{4}~~~
Dahun Kim\textsuperscript{3}~~~
Qihang Yu\textsuperscript{2}~~~
Daniel Cremers\textsuperscript{1}~~~\\
Laura Leal-Taix\'{e}\textsuperscript{1}~~~
Alan L. Yuille\textsuperscript{2}~~~
Florian Schroff\textsuperscript{4}~~~
Hartwig Adam\textsuperscript{4}~~~
Liang-Chieh Chen\textsuperscript{4}\\
\textsuperscript{1}Technical University Munich~~~
\textsuperscript{2}Johns Hopkins University~~~
\textsuperscript{3}KAIST~~~
\textsuperscript{4}Google Research
}

\maketitle

\input{sections/0.abstract.tex}
\input{sections/1.intro.tex}
\input{sections/2.task.tex}
\input{sections/3.model.tex}
\input{sections/4.backbone.tex}

\input{sections/5.training.tex}
\input{sections/6.conclusion.tex}

{\small
\bibliographystyle{ieee_fullname}
\bibliography{refs}
}

\end{document}

%% file: sections/0.abstract.tex
\begin{abstract}
DeepLab2 is a TensorFlow library for deep labeling, aiming to provide a state-of-the-art and easy-to-use TensorFlow codebase for general dense pixel prediction problems in computer vision. DeepLab2 includes all our recently developed DeepLab model variants with pretrained checkpoints as well as model training and evaluation code, allowing the community to reproduce and further improve upon the state-of-art systems. To showcase the effectiveness of DeepLab2, our Panoptic-DeepLab employing Axial-SWideRNet as network backbone achieves 68.0\% PQ or 83.5\% mIoU on Cityscaspes validation set, with only single-scale inference and ImageNet-1K pretrained checkpoints. We hope that publicly sharing our library could facilitate future research on dense pixel labeling tasks and envision new applications of this technology. Code is made publicly available at \url{https://github.com/google-research/deeplab2}.
\end{abstract}

%% file: sections/1.intro.tex
\section{Introduction}
Deep labeling refers to solving certain computer vision problems by assigning a predicted value for each pixel (\ie, label each pixel) in an image or video with a deep neural network~\cite{lecun1998gradient,long2014fully,deeplabv12015}. Typical dense prediction problems  include, but not limited to, semantic segmentation~\cite{he2004multiscale,ladicky2009associative,everingham2010pascal}, instance segmentation~\cite{hariharan2014simultaneous,lin2014microsoft}, panoptic segmentation~\cite{kirillov2018panoptic,neuhold2017mapillary}, depth estimation~\cite{Silberman:ECCV12,geiger2013vision}, video panoptic segmentation~\cite{kim2020video,weber2021step}, and depth-aware video panoptic segmentation~\cite{qiao2021vip}.

Going beyond our previous open source library\footnote{\url{https://github.com/tensorflow/models/tree/master/research/deeplab}} in 2018 (which could only tackle image semantic segmentation with the first few DeepLab model variants~\cite{deeplabv12015,chen2018deeplabv2,chen2017deeplabv3,deeplabv3plus2018}), we introduce DeepLab2, a modern TensorFlow library~\cite{tensorflow-osdi2016} for deep labeling, aiming to provide a unified and easy-to-use TensorFlow codebase for general dense pixel labeling tasks.
\textit{Re-implemented} in TensorFlow2, this release includes \textit{all} our recently developed DeepLab model variants~\cite{cheng2019panoptic,wang2020axial,wang2021max,weber2021step,qiao2021vip}, model training and evaluation code, and several pretrained checkpoints, allowing the community to reproduce and further improve upon the state-of-art systems. We hope that the open source DeepLab2 would facilitate future research on dense pixel labeling tasks, and anticipate novel breakthroughs and new applications that adopt this technology.

In the following sections, we detail a few popular dense prediction tasks as well as the provided state-of-the-art models in the DeepLab2 library.

%% file: sections/2.task.tex
\section{Dense Pixel Labeling Tasks}

Several computer vision problems could be formulated as dense pixel labeling. In this section, we briefly introduce some typical examples of dense pixel labeling tasks.

\begin{figure*}[!th]
\centering
    \begin{tabular}{cccc}
         \includegraphics[height=0.11\linewidth,width=0.22\linewidth]{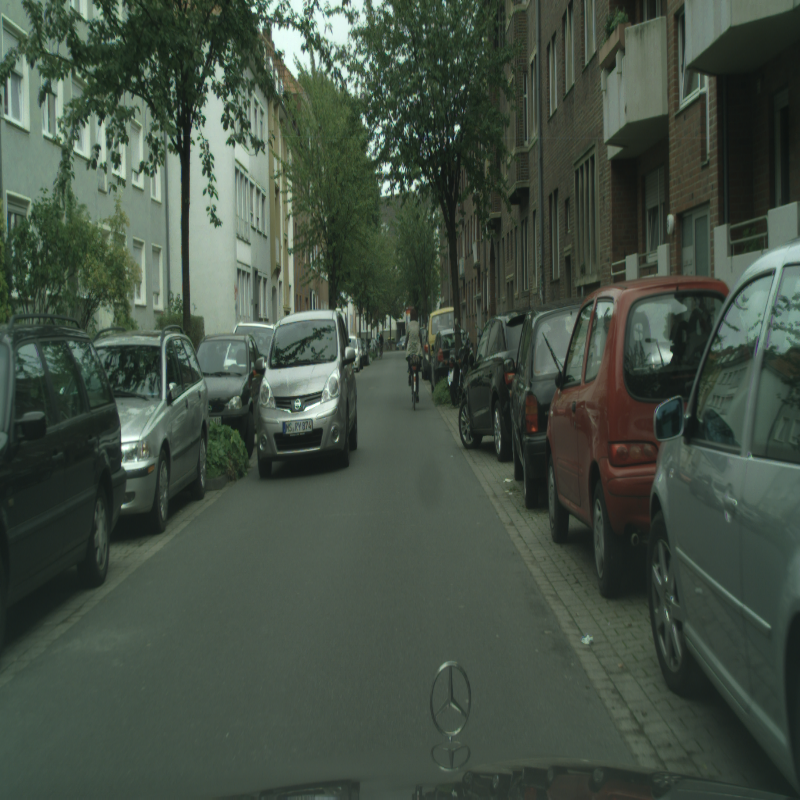} &
         \includegraphics[height=0.11\linewidth,width=0.22\linewidth]{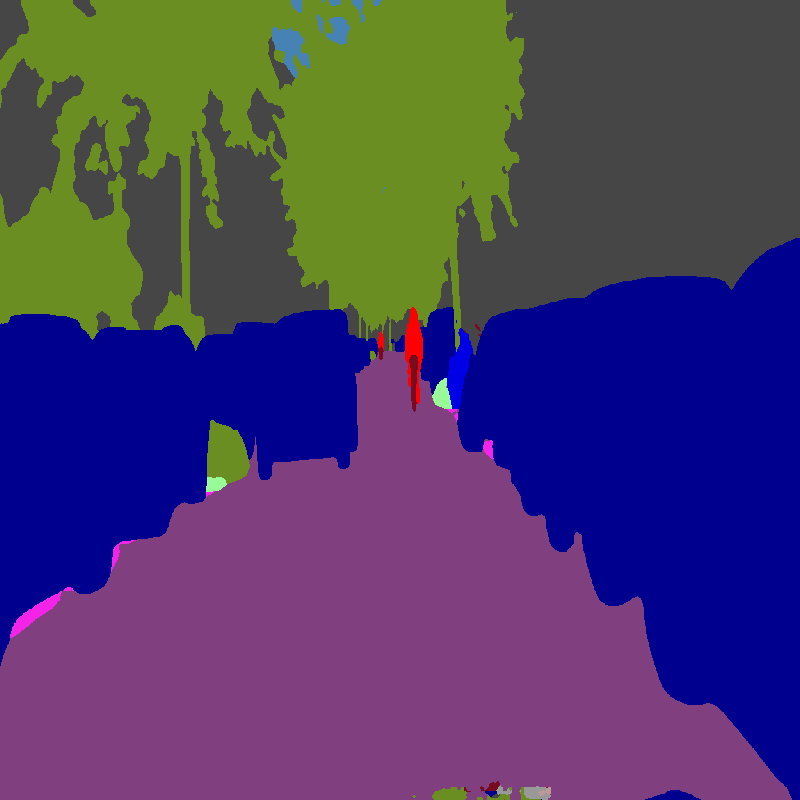} &
         \includegraphics[height=0.11\linewidth,width=0.22\linewidth]{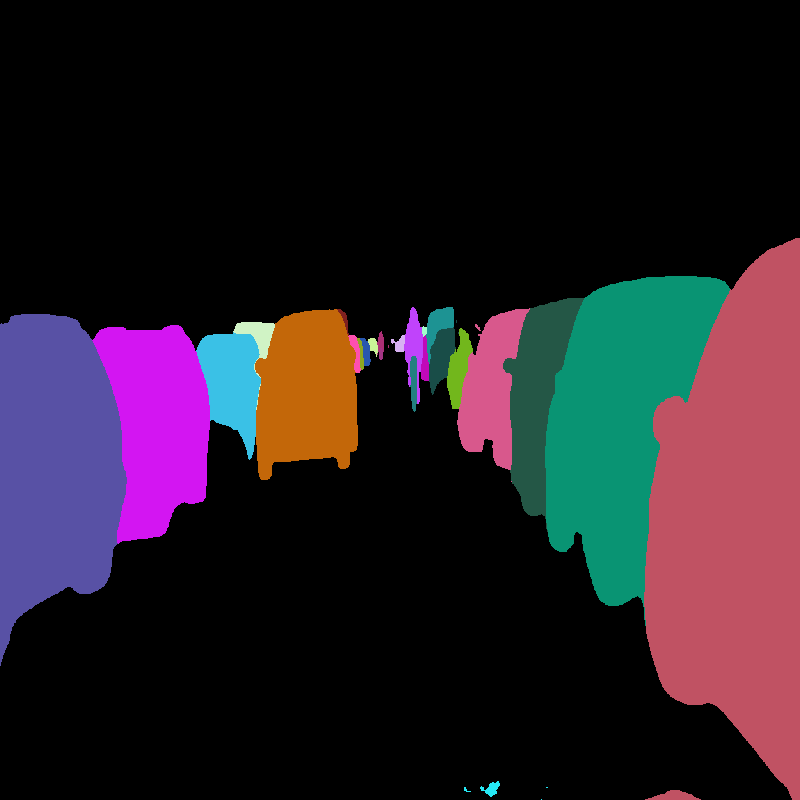} &
         \includegraphics[height=0.11\linewidth,width=0.22\linewidth]{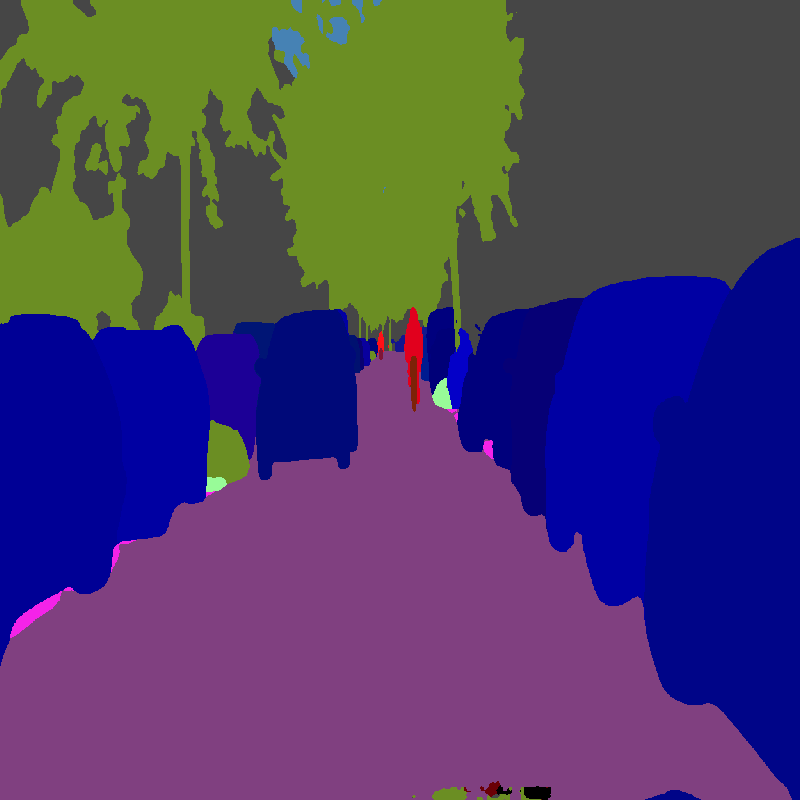} \\
         (a) Image & (b) Semantic Segmentation & (c) Instance Segmentation & (d) Panoptic Segmentation \\
    \end{tabular}
    \caption{Segmentation results by Panoptic-DeepLab~\cite{cheng2019panoptic} on Cityscapes Images~\cite{cordts2016cityscapes}.}
    \label{fig:image_seg}
\end{figure*}

\begin{figure*}[!th]
\centering
\begin{subfigure}{.245\linewidth}
  \centering
  \includegraphics[trim={0 100 0 100},clip,width=\linewidth]{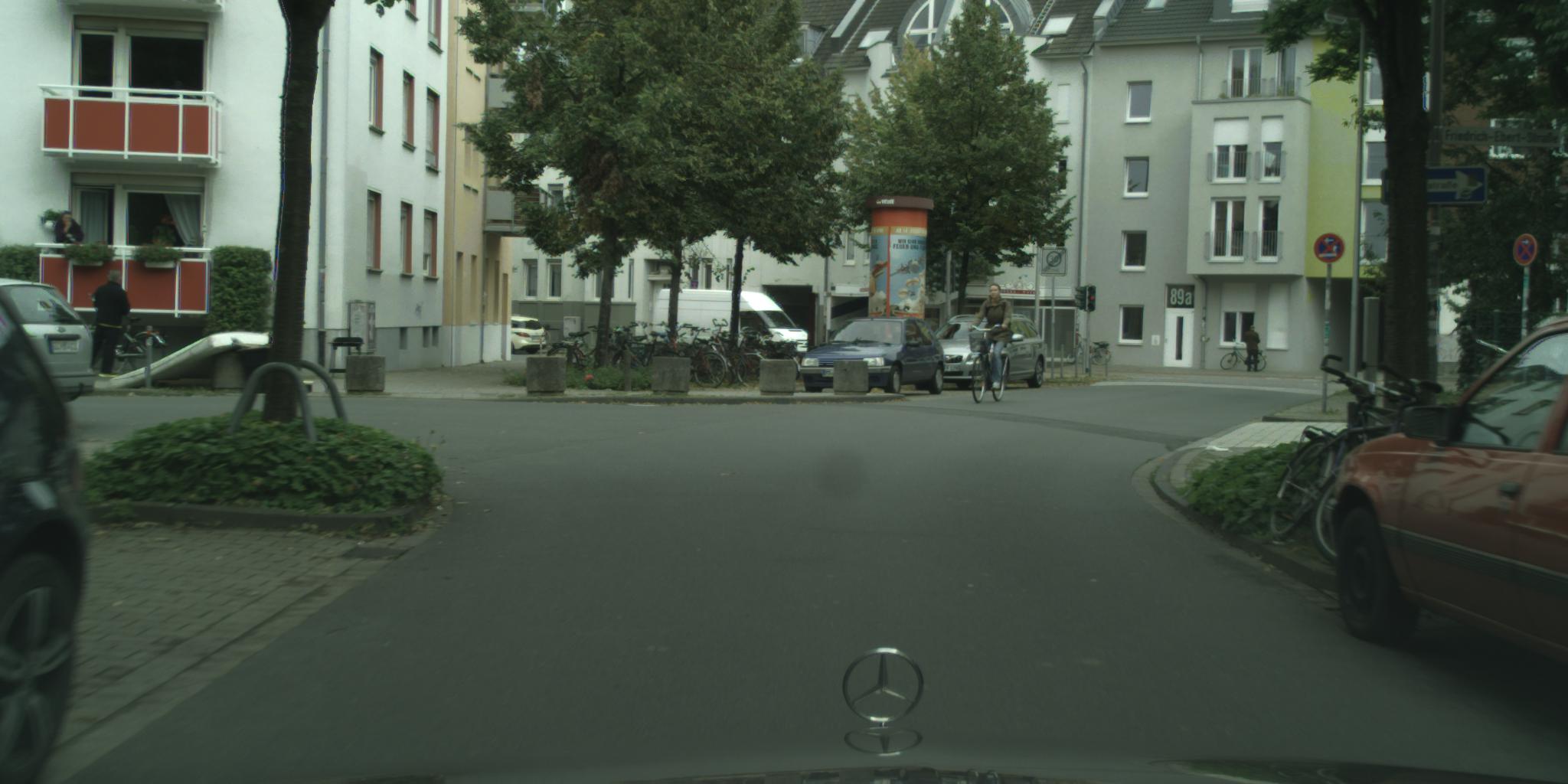}
\end{subfigure}\hfill\begin{subfigure}{.245\linewidth}
  \centering
  \includegraphics[trim={0 100 0 100},clip,width=\linewidth]{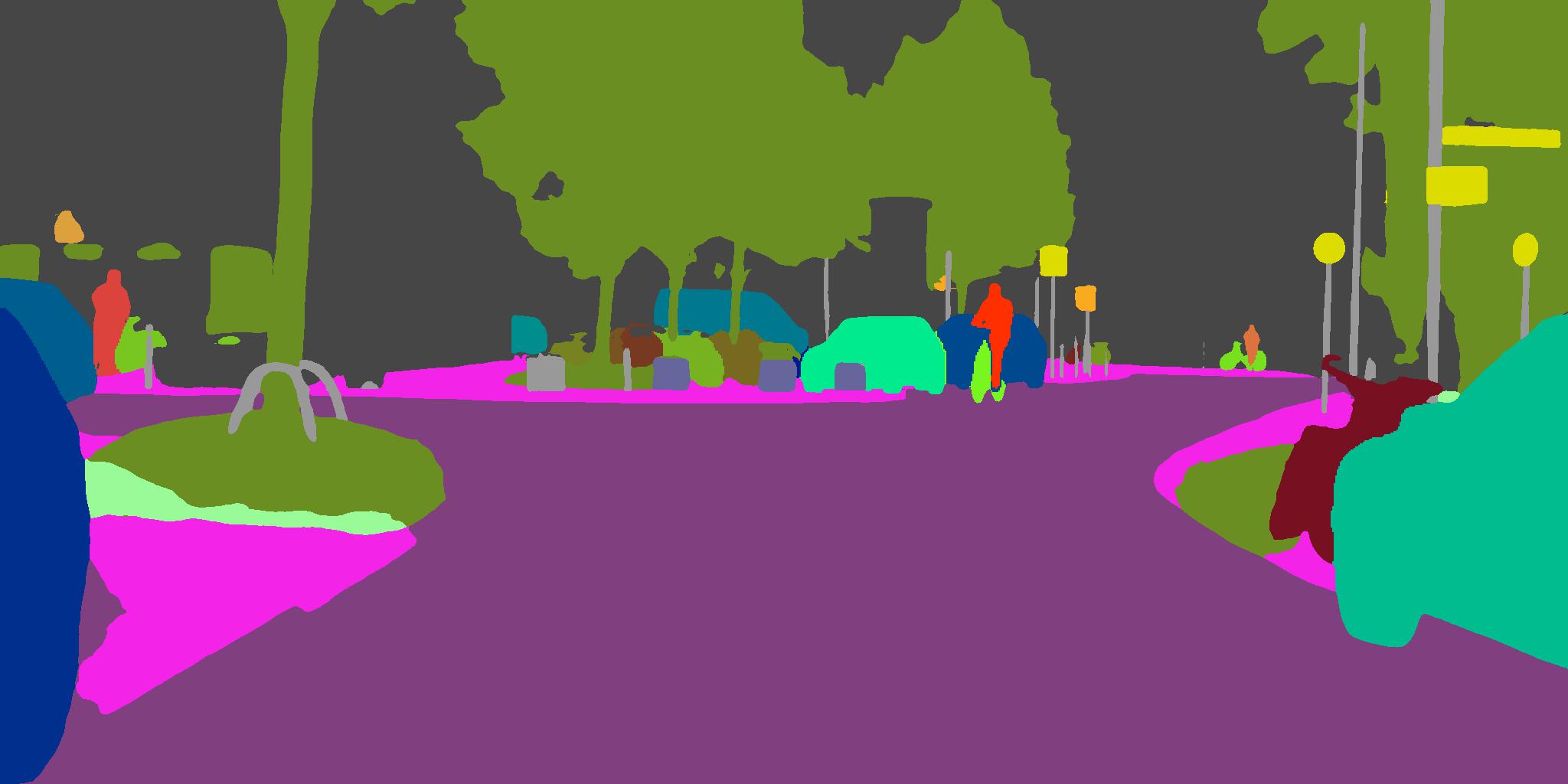}
\end{subfigure}\hfill\begin{subfigure}{.245\linewidth}
  \centering
  \includegraphics[trim={0 100 0 100},clip,width=\linewidth]{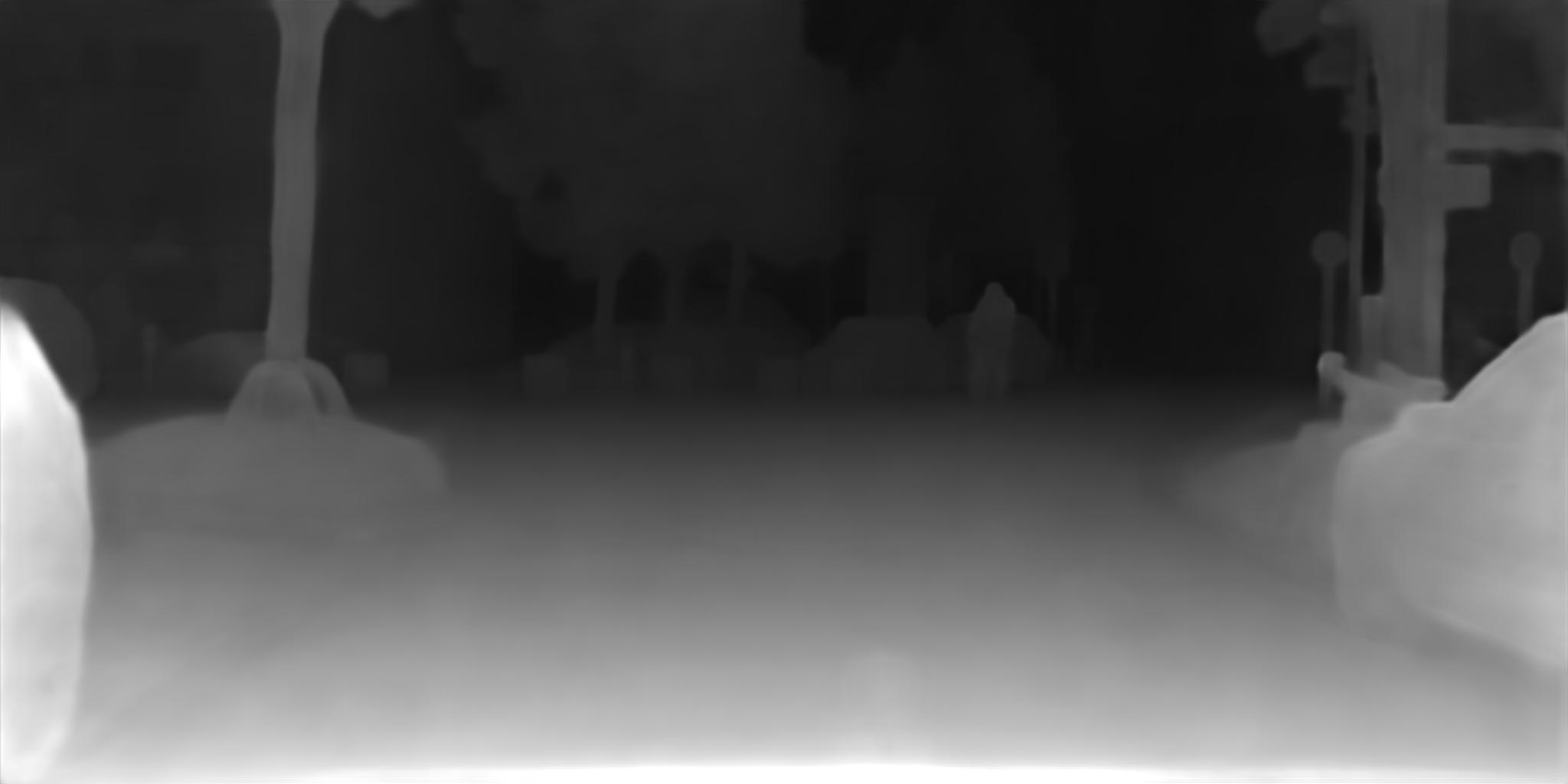}
\end{subfigure}\hfill\begin{subfigure}{.245\linewidth}
  \centering
  \includegraphics[trim={400 130 0 200},clip,width=\linewidth]{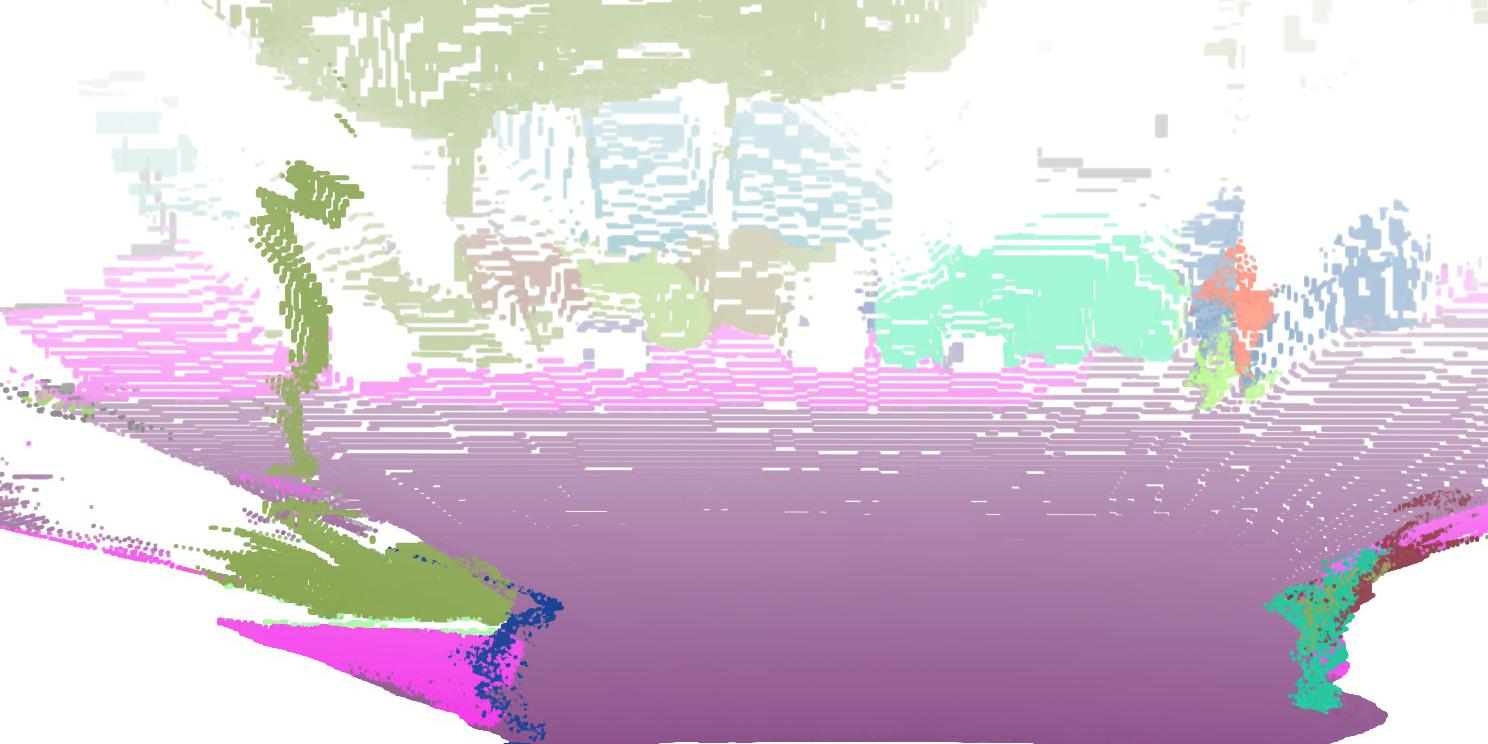}
\end{subfigure}\\\begin{subfigure}{.245\linewidth}
  \centering
  \includegraphics[trim={0 100 0 100},clip,width=\linewidth]{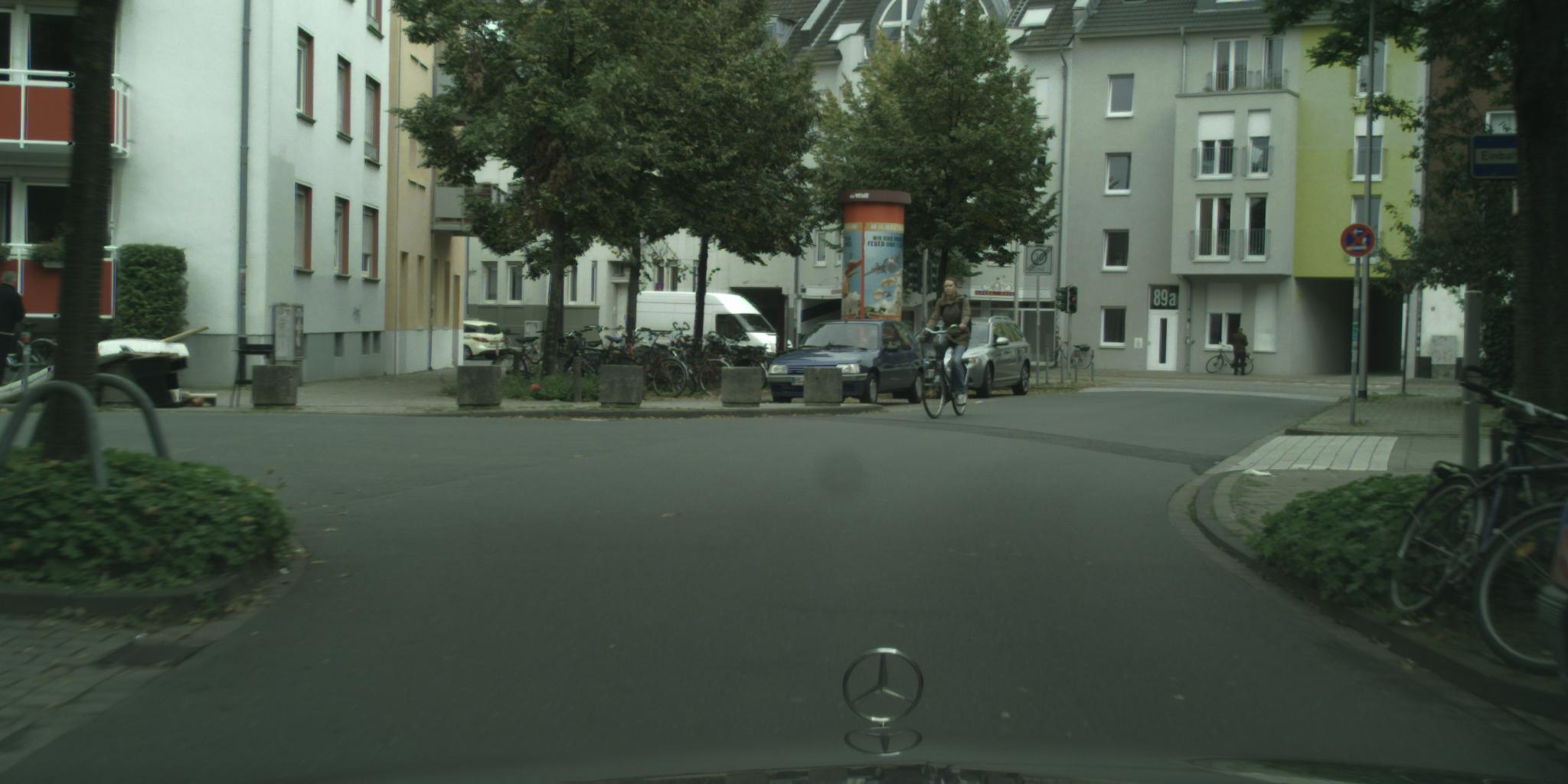}
\end{subfigure}\hfill\begin{subfigure}{.245\linewidth}
  \centering
  \includegraphics[trim={0 100 0 100},clip,width=\linewidth]{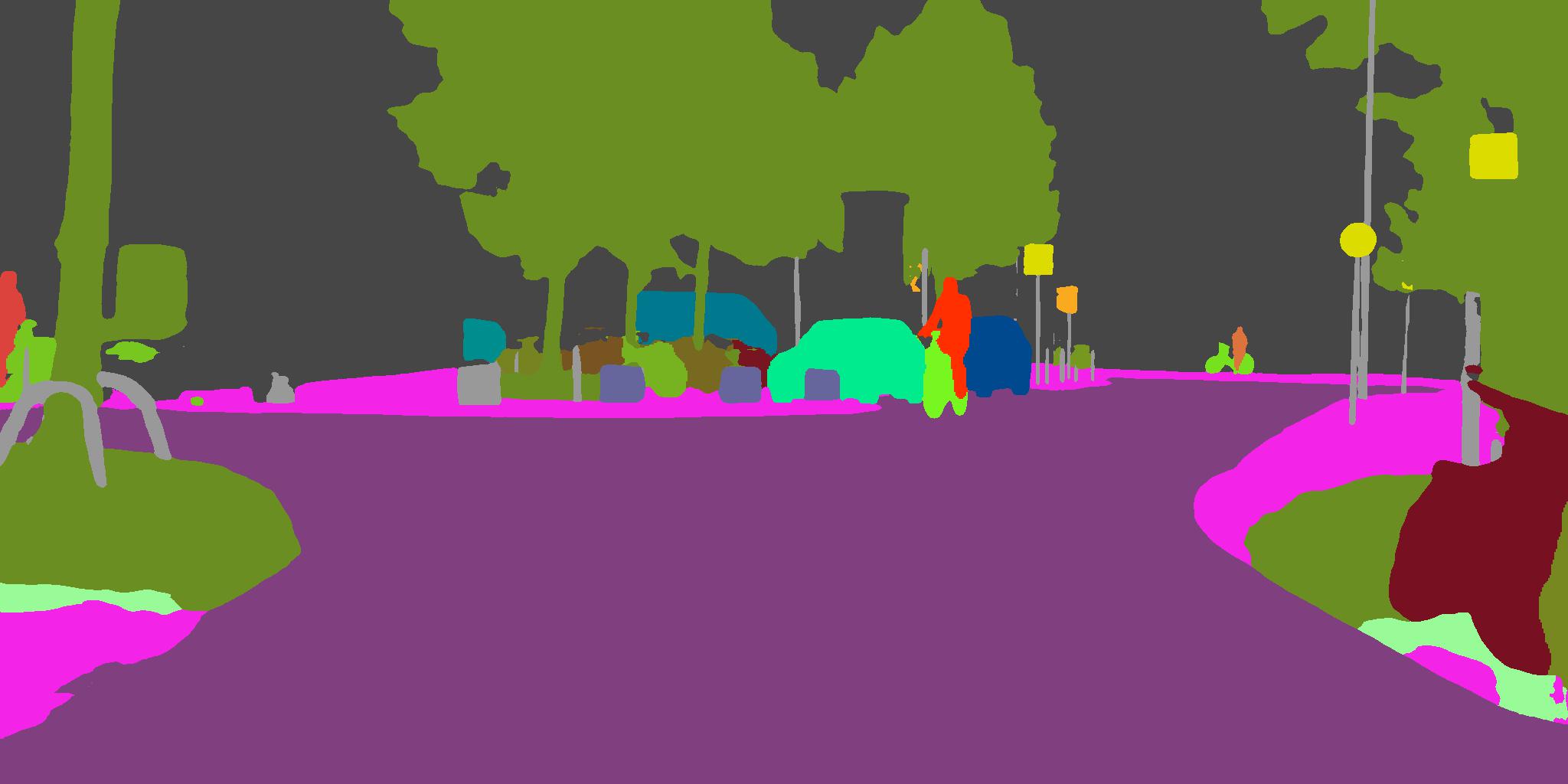}
\end{subfigure}\hfill\begin{subfigure}{.245\linewidth}
  \centering
  \includegraphics[trim={0 100 0 100},clip,width=\linewidth]{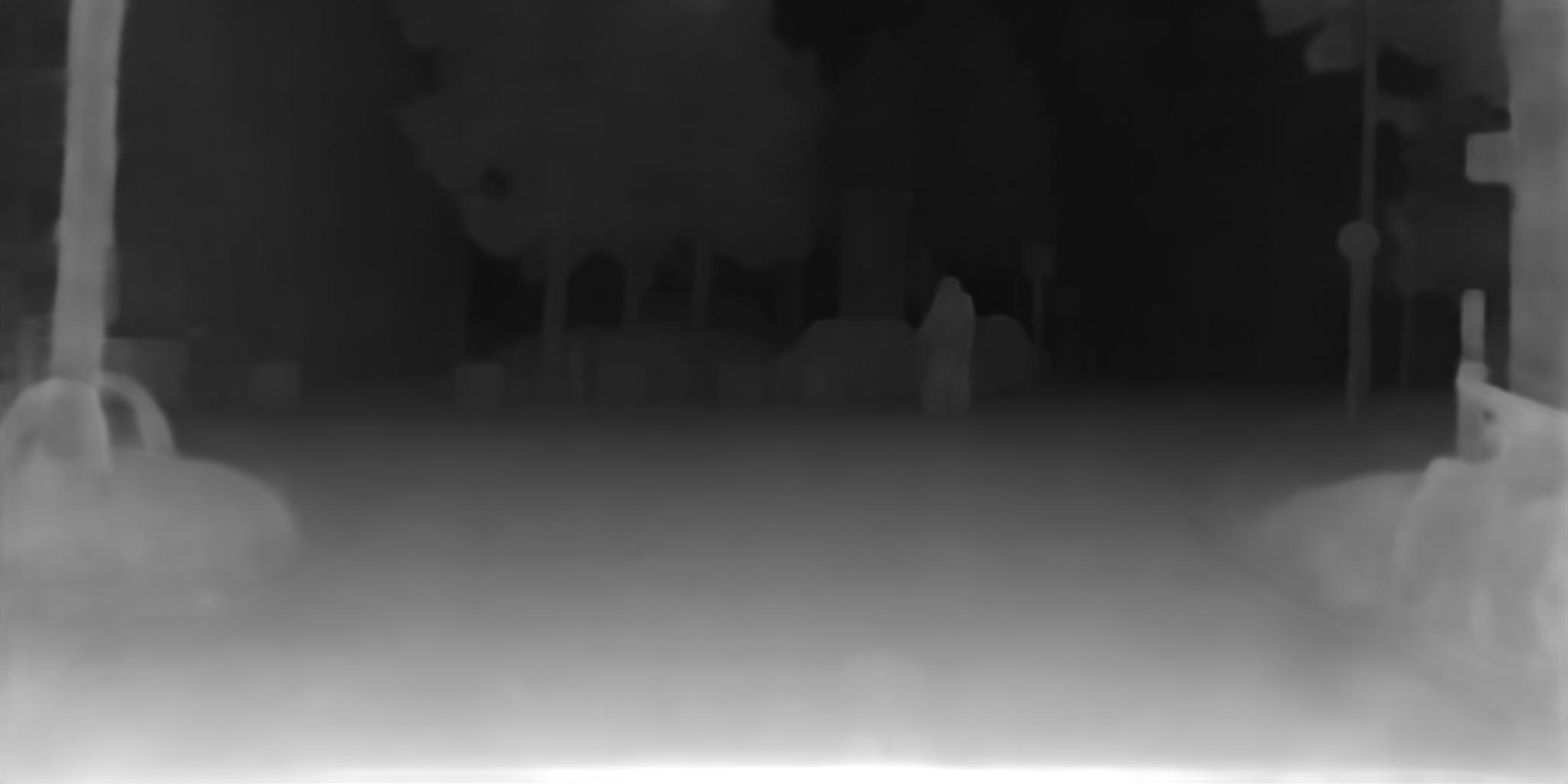}
\end{subfigure}\hfill\begin{subfigure}{.245\linewidth}
  \centering
  \includegraphics[trim={400 130 0 200},clip,width=\linewidth]{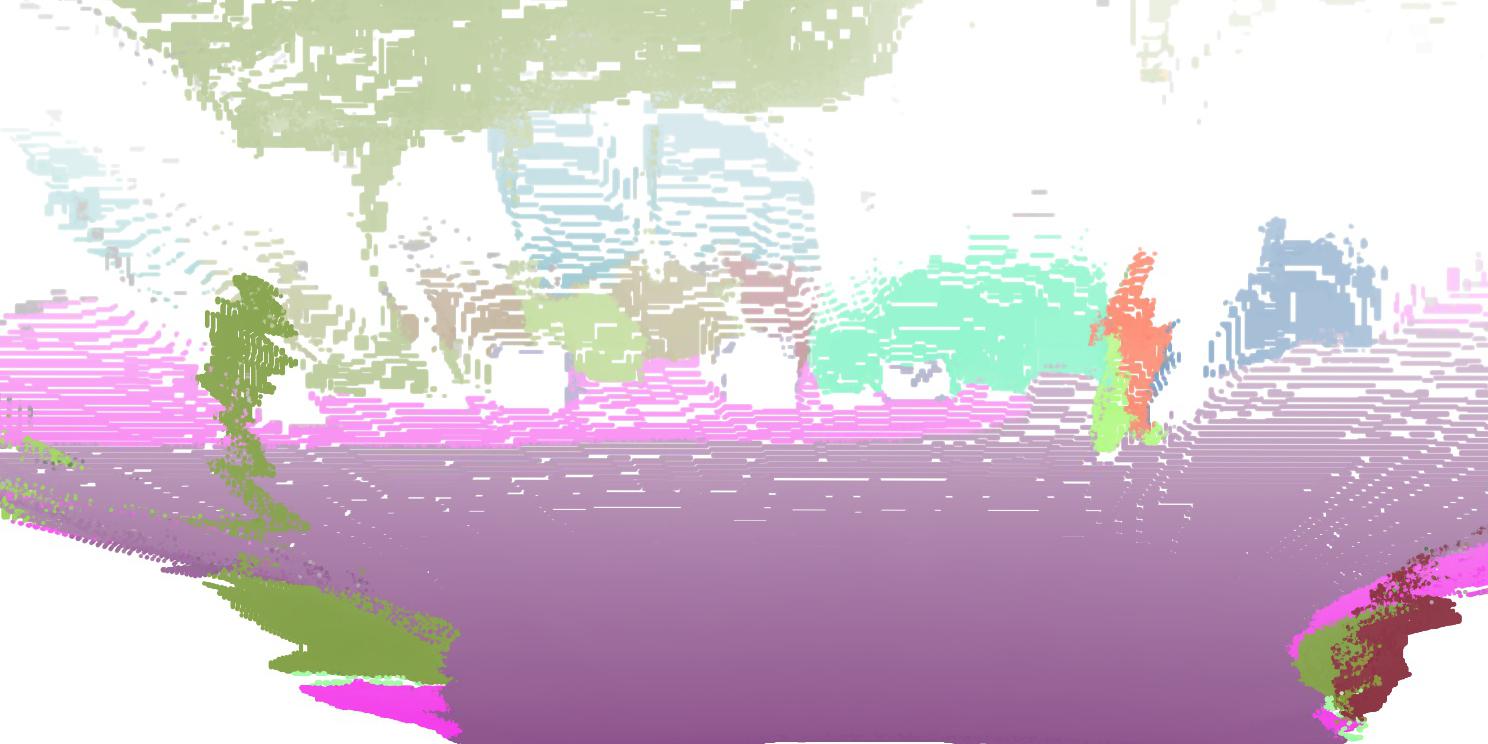}
\end{subfigure}\\\begin{subfigure}{.245\linewidth}
  \centering
  \includegraphics[trim={0 100 0 100},clip,width=\linewidth]{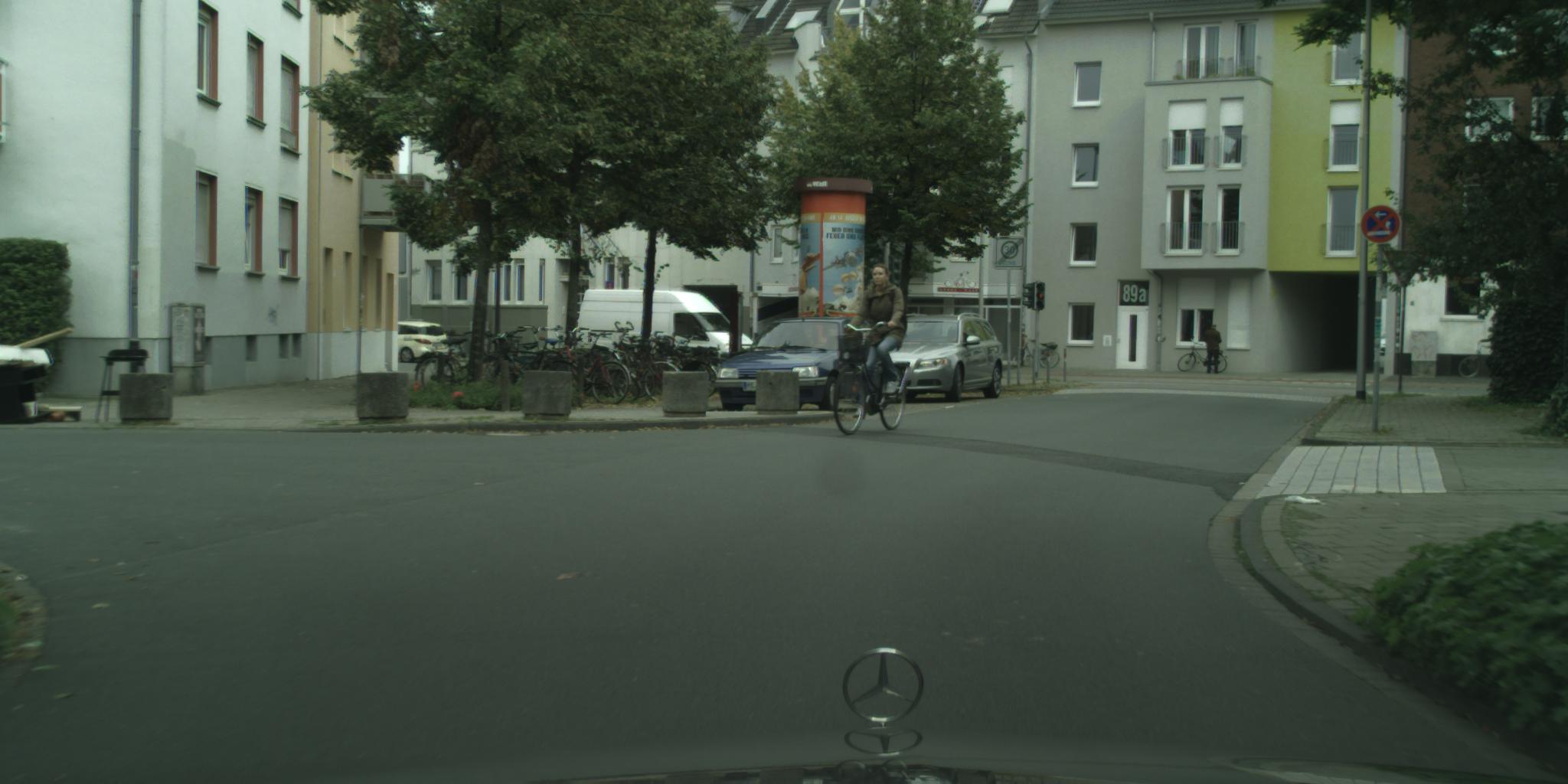}
\end{subfigure}\hfill\begin{subfigure}{.245\linewidth}
  \centering
  \includegraphics[trim={0 100 0 100},clip,width=\linewidth]{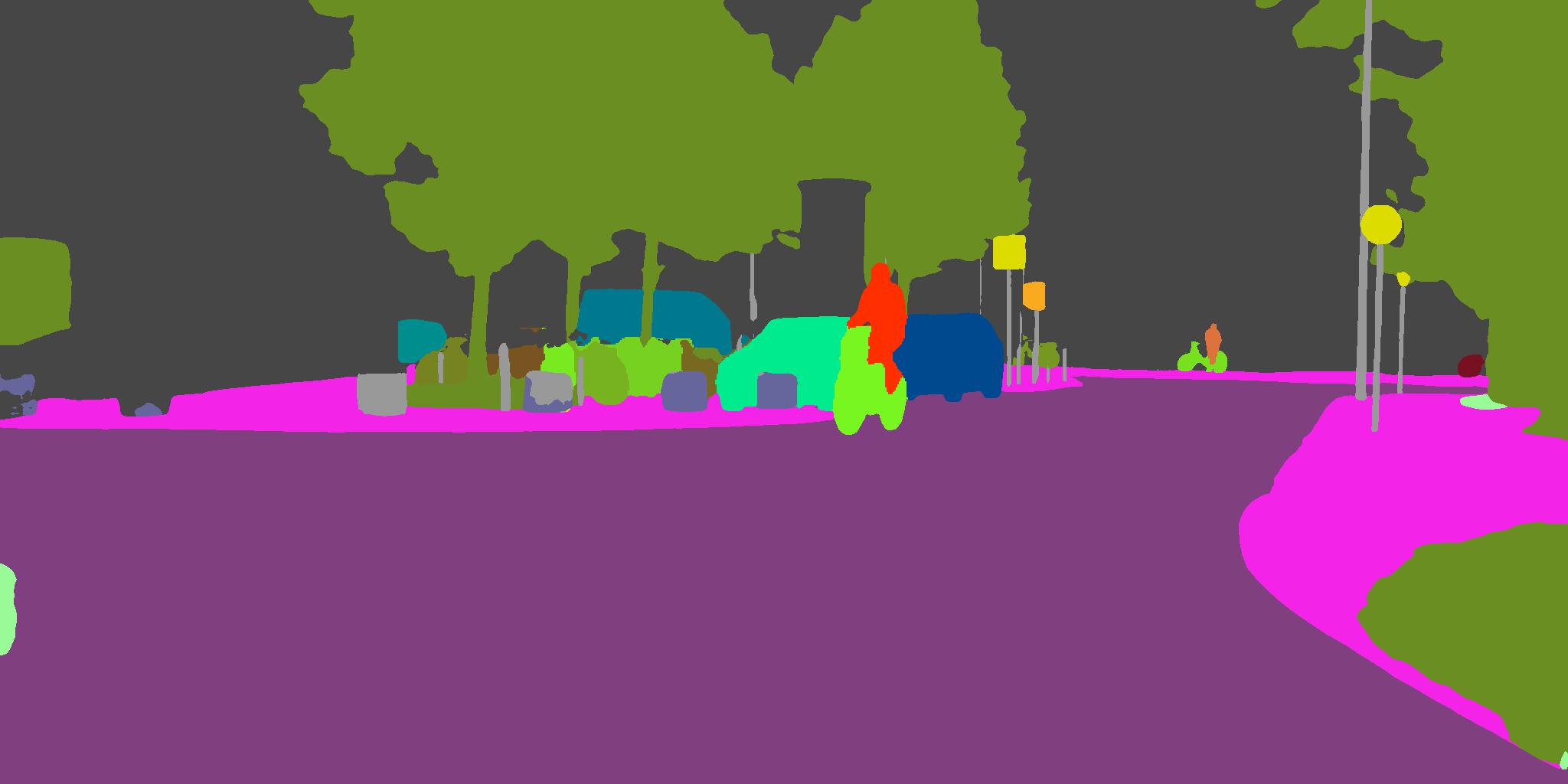}
\end{subfigure}\hfill\begin{subfigure}{.245\linewidth}
  \centering
  \includegraphics[trim={0 100 0 100},clip,width=\linewidth]{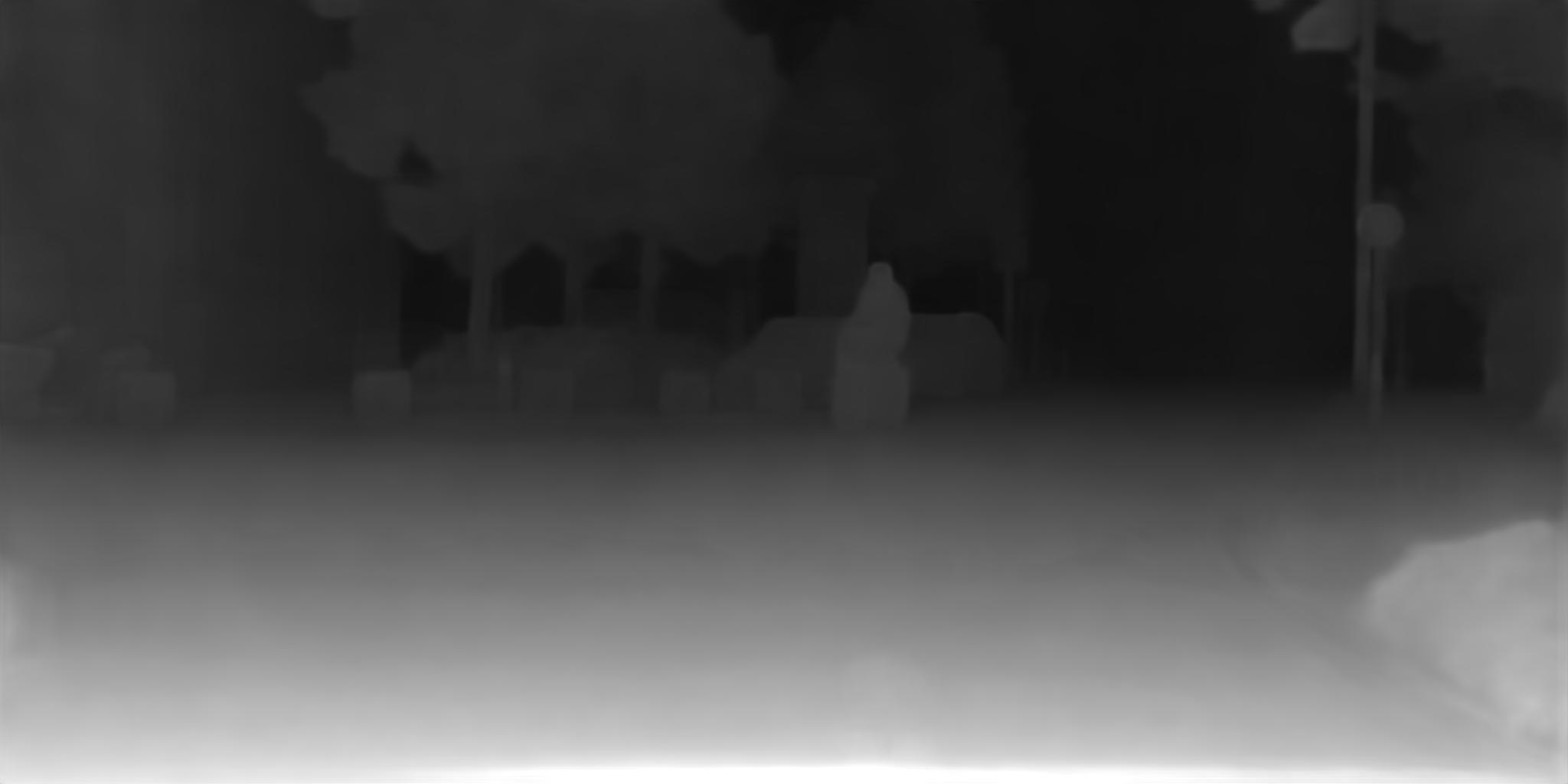}
\end{subfigure}\hfill\begin{subfigure}{.245\linewidth}
  \centering
  \includegraphics[trim={400 130 0 200},clip,width=\linewidth]{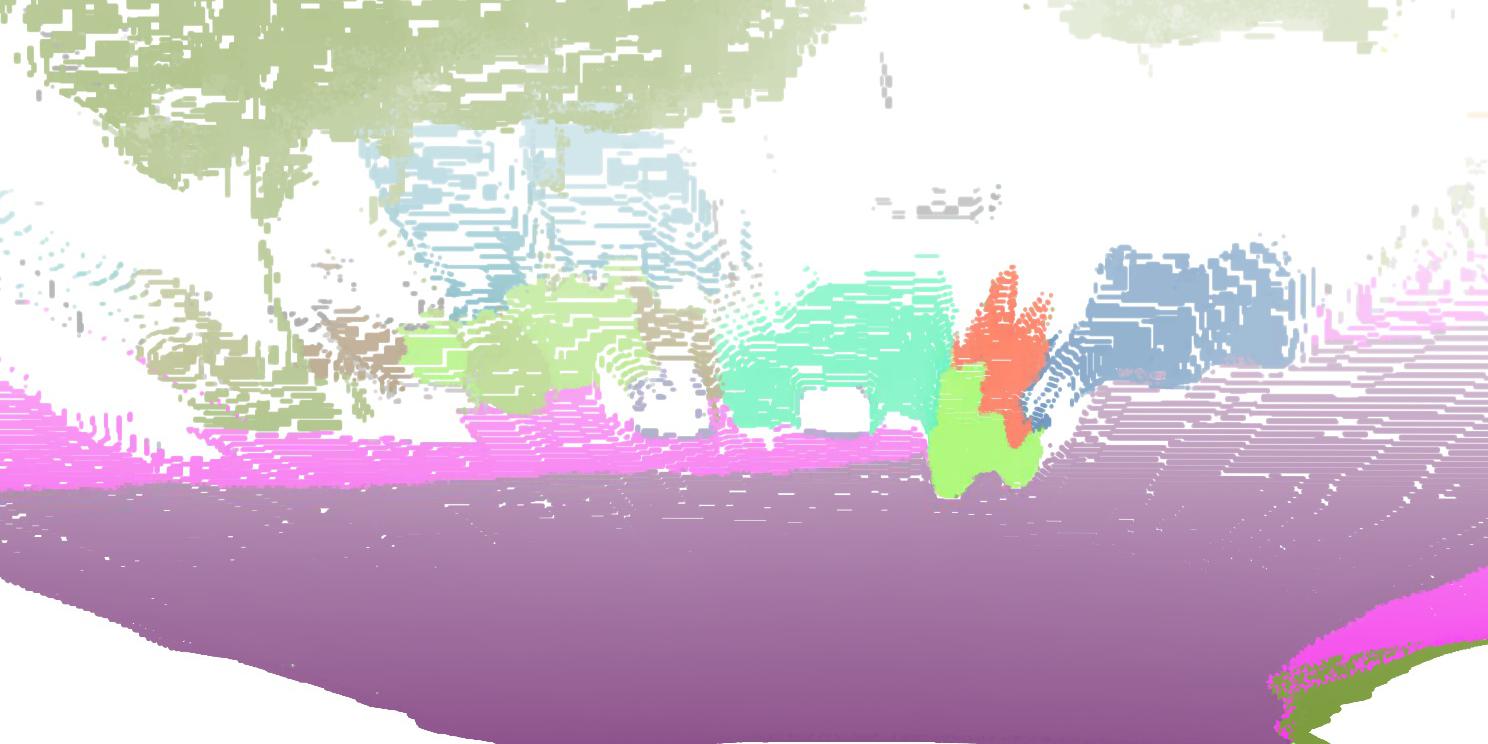}
\end{subfigure}\\\begin{subfigure}{.245\linewidth}
  \centering
  \includegraphics[trim={0 100 0 100},clip,width=\linewidth]{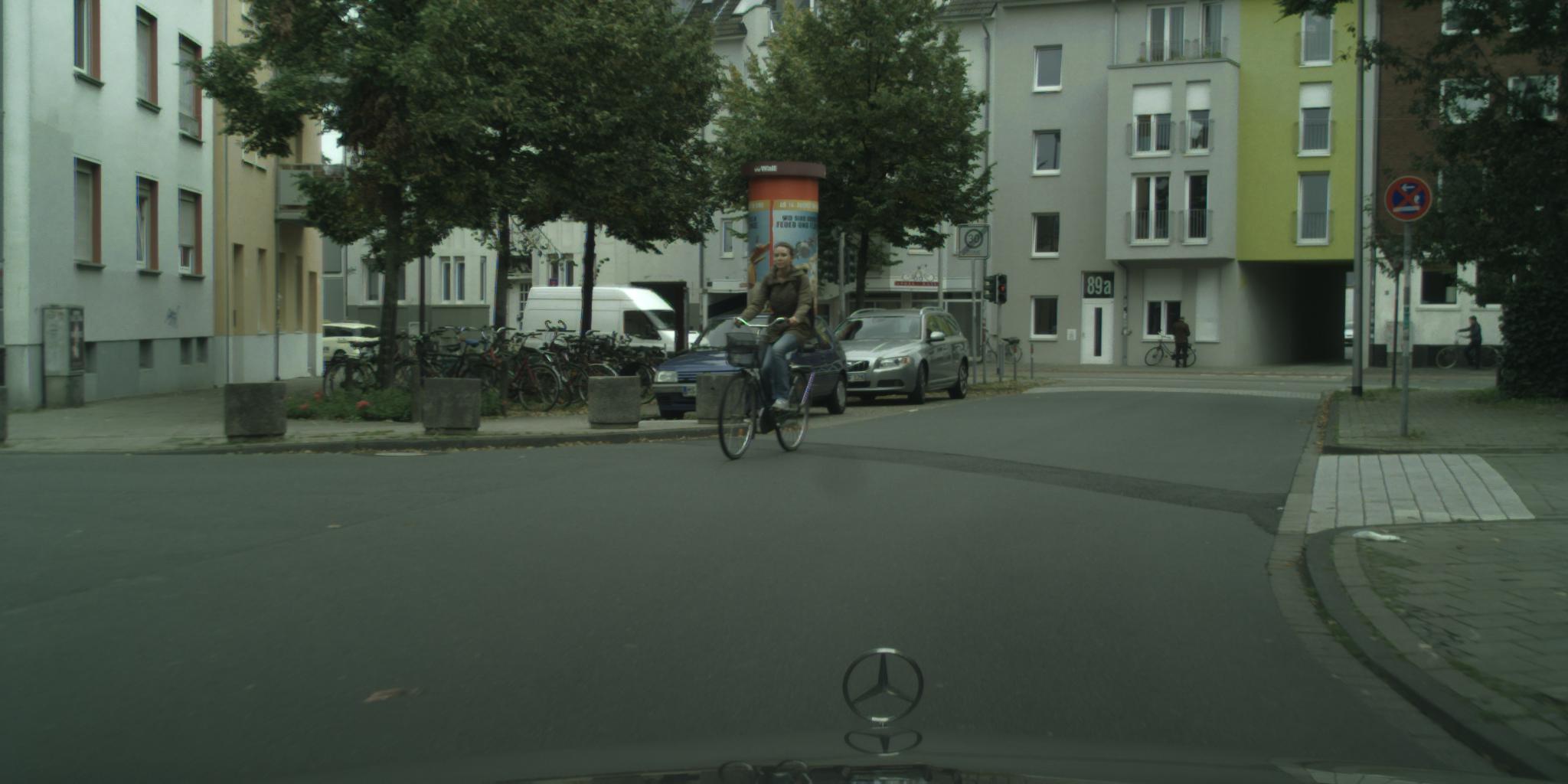}
\end{subfigure}\hfill\begin{subfigure}{.245\linewidth}
  \centering
  \includegraphics[trim={0 100 0 100},clip,width=\linewidth]{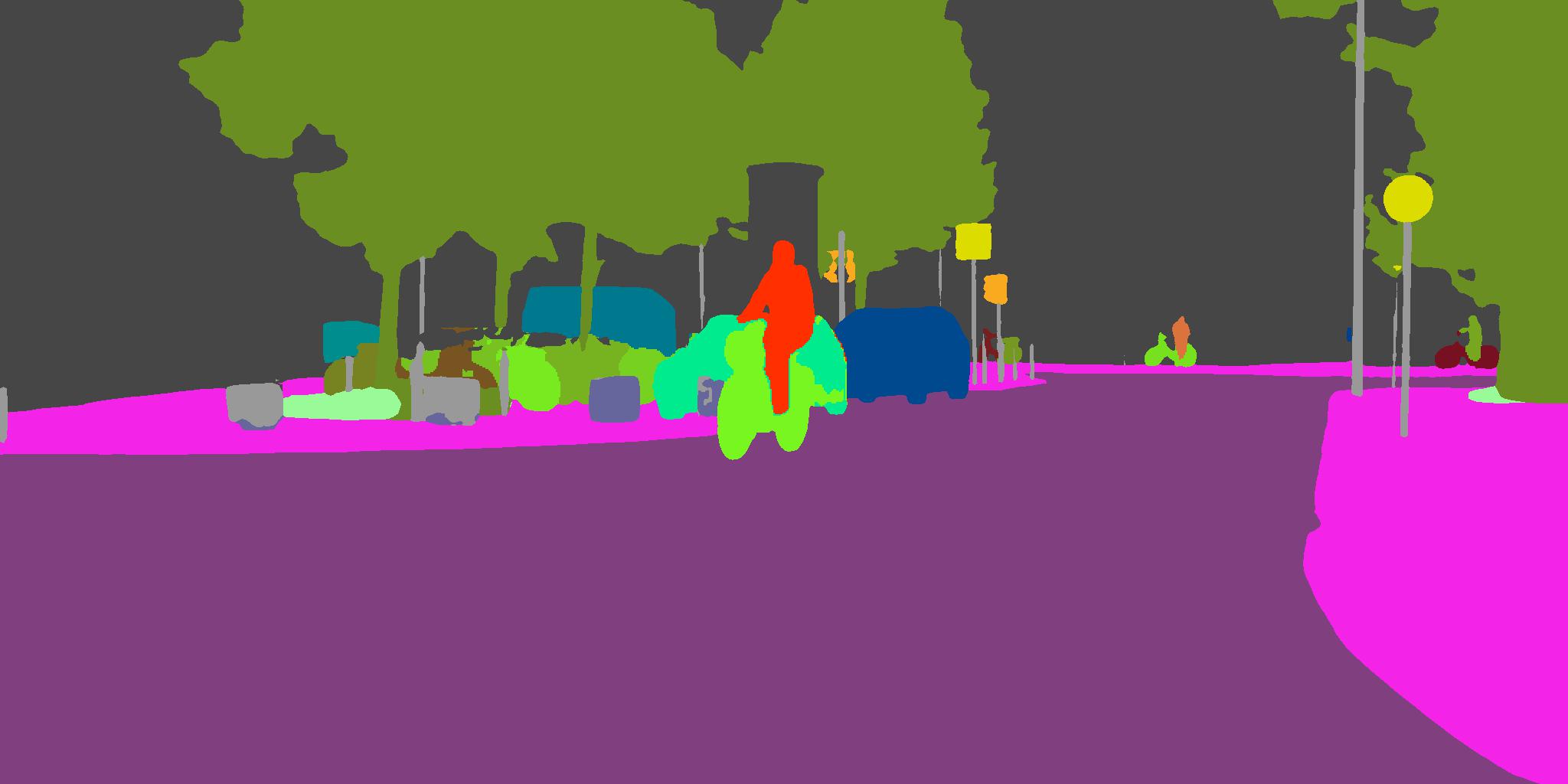}
\end{subfigure}\hfill\begin{subfigure}{.245\linewidth}
  \centering
  \includegraphics[trim={0 100 0 100},clip,width=\linewidth]{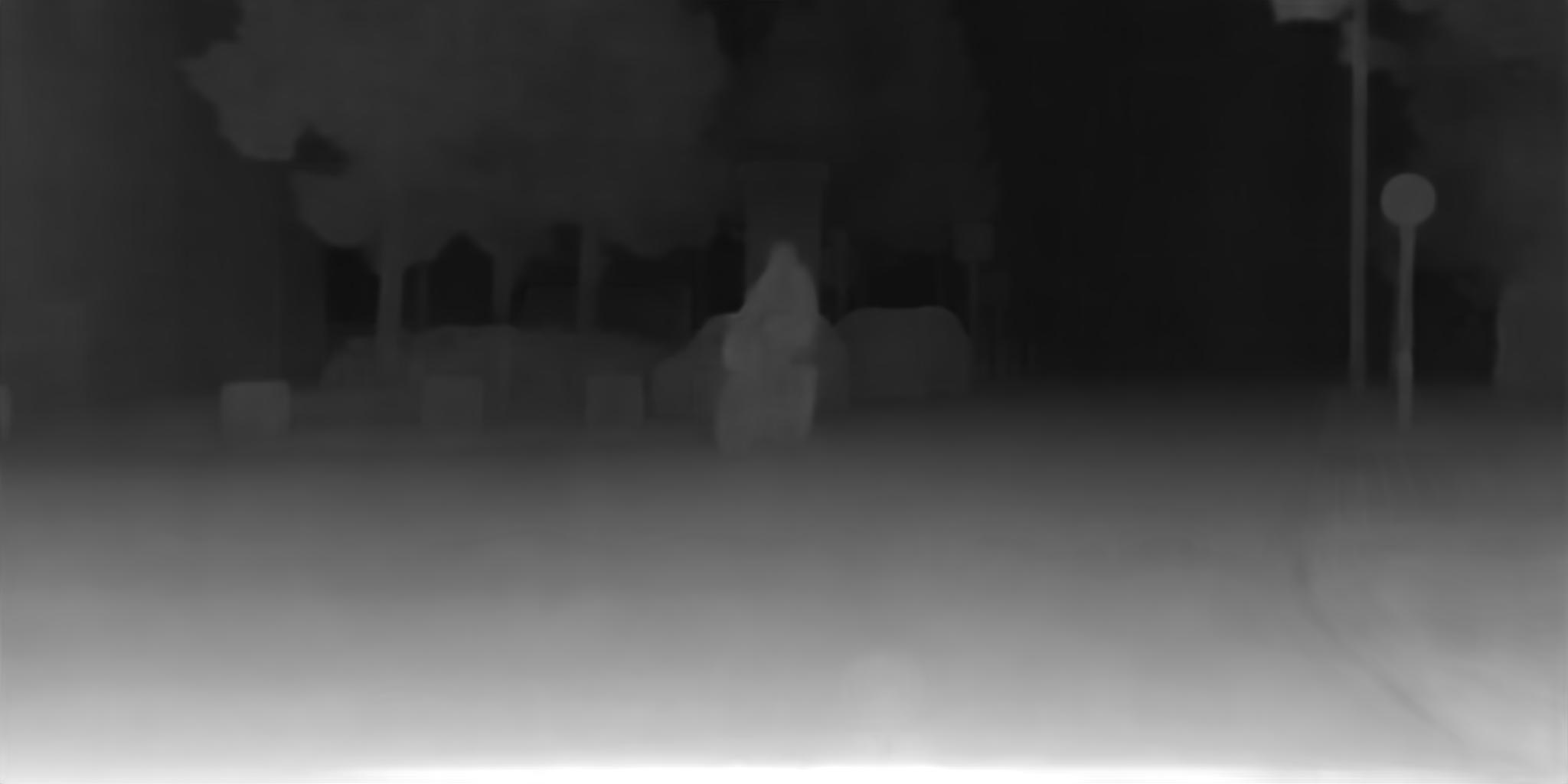}
\end{subfigure}\hfill\begin{subfigure}{.245\linewidth}
  \centering
  \includegraphics[trim={400 130 0 200},clip,width=\linewidth]{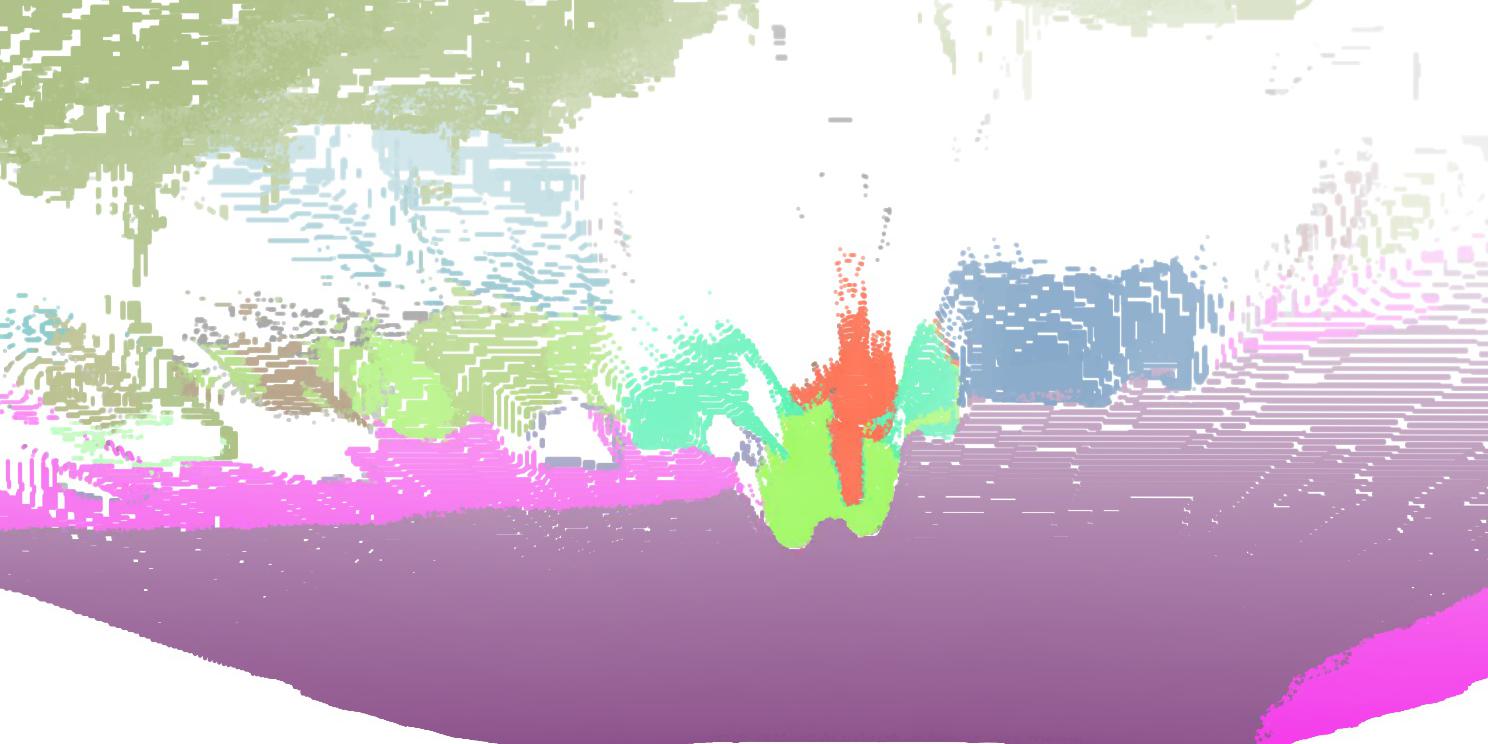}
\end{subfigure}\\\begin{subfigure}{.245\linewidth}
  \centering
  \includegraphics[trim={0 100 0 100},clip,width=\linewidth]{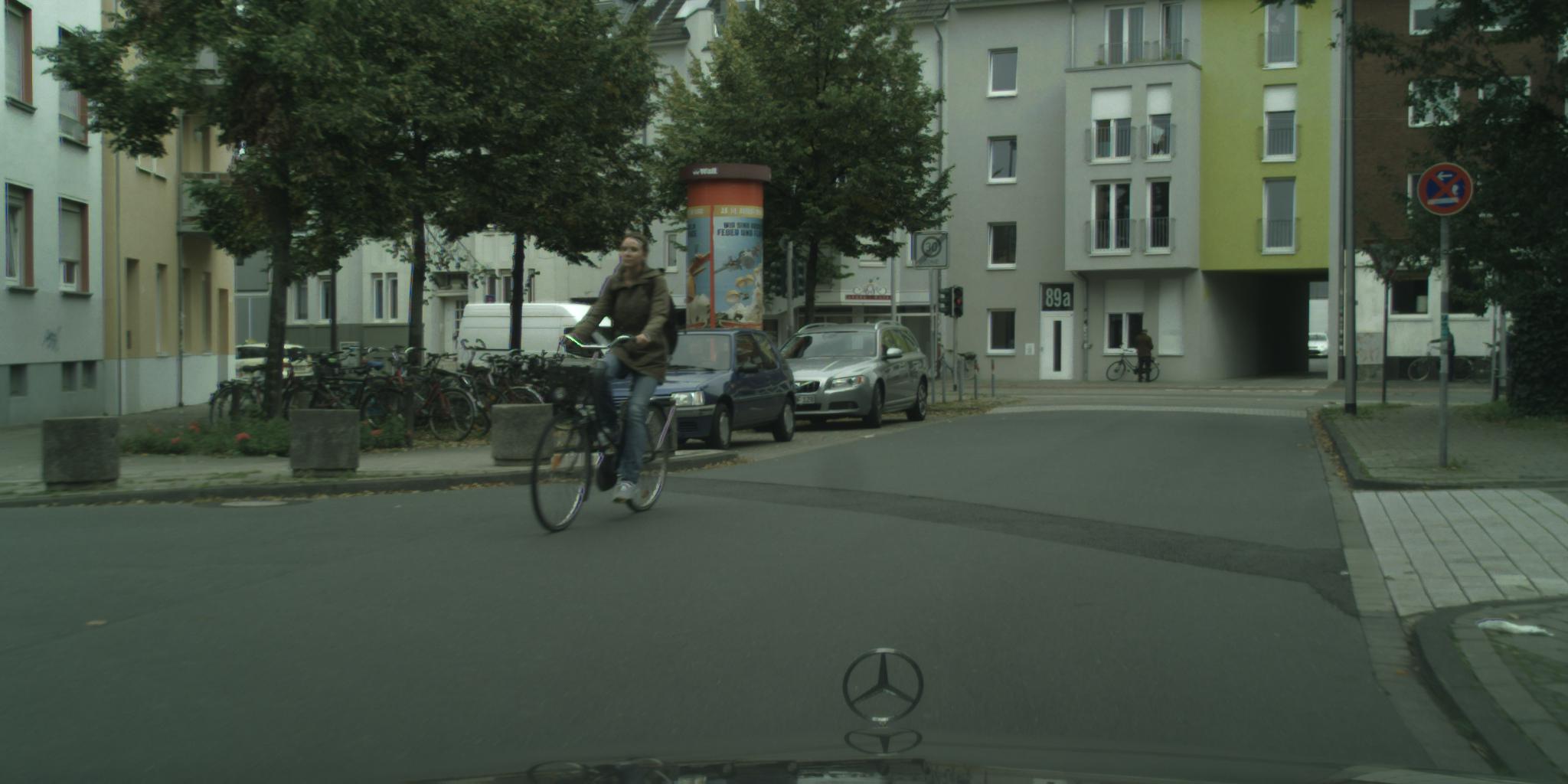}
\end{subfigure}\hfill\begin{subfigure}{.245\linewidth}
  \centering
  \includegraphics[trim={0 100 0 100},clip,width=\linewidth]{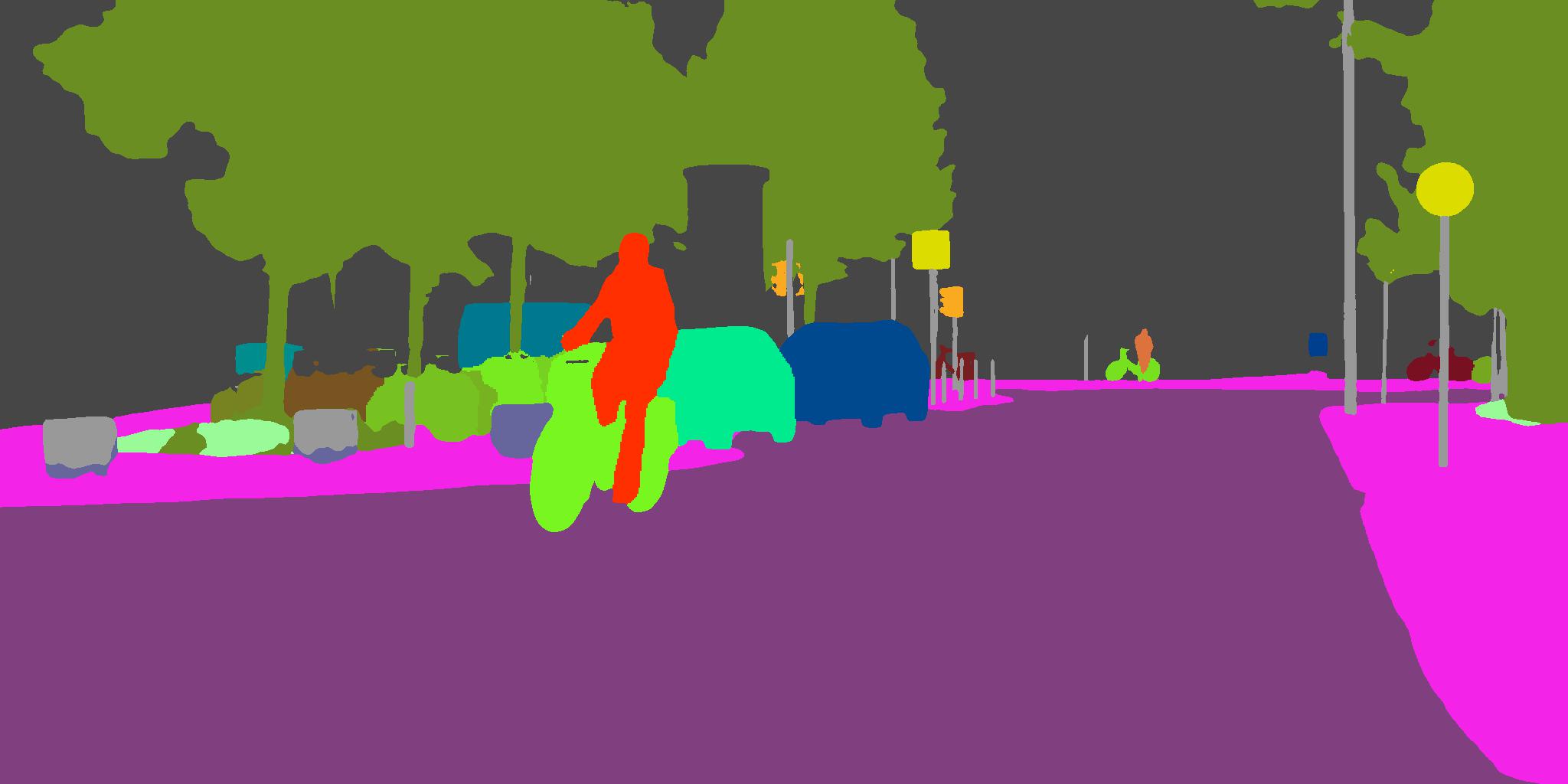}
\end{subfigure}\hfill\begin{subfigure}{.245\linewidth}
  \centering
  \includegraphics[trim={0 100 0 100},clip,width=\linewidth]{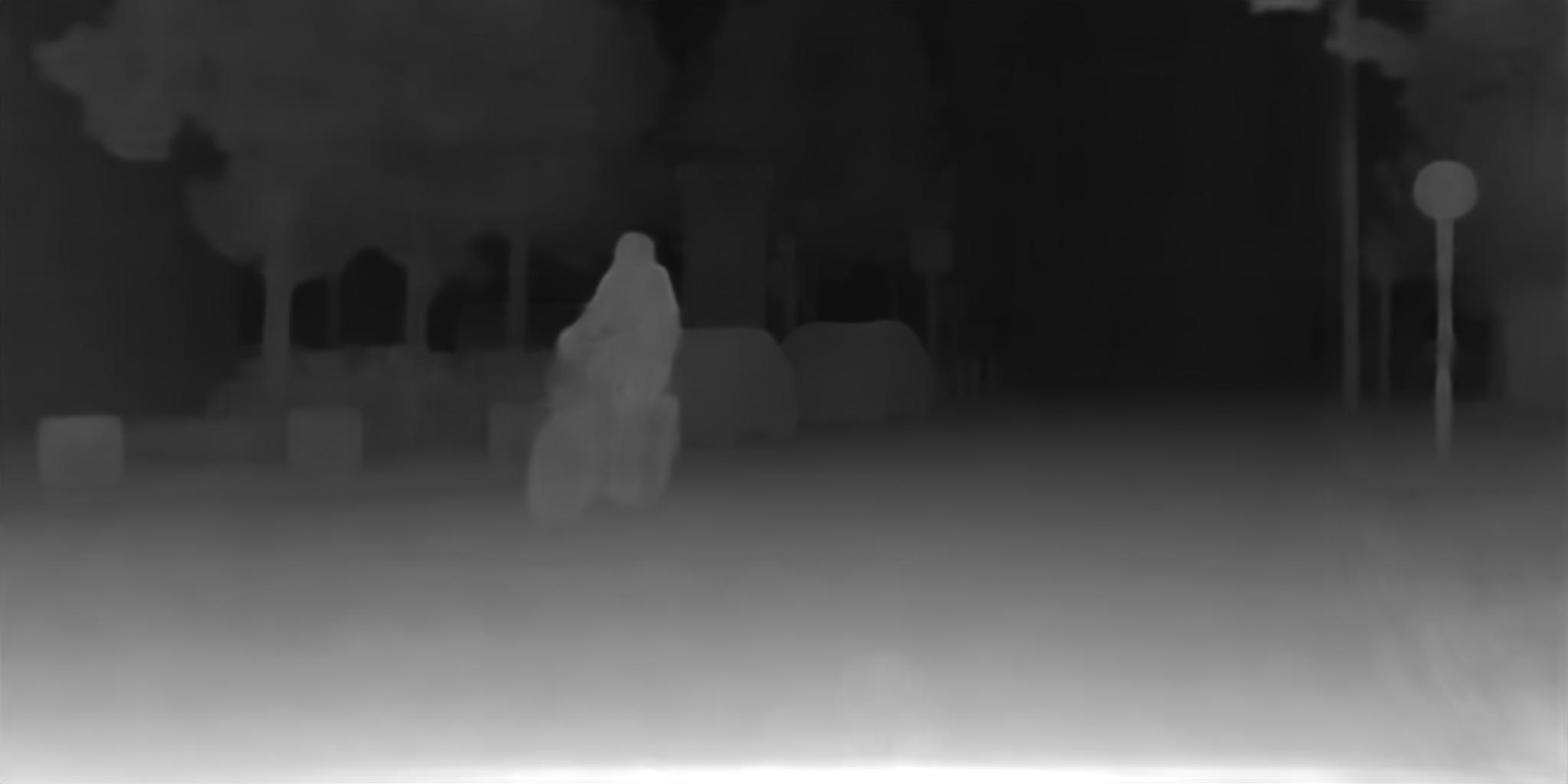}
\end{subfigure}\hfill\begin{subfigure}{.245\linewidth}
  \centering
  \includegraphics[trim={400 130 0 200},clip,width=\linewidth]{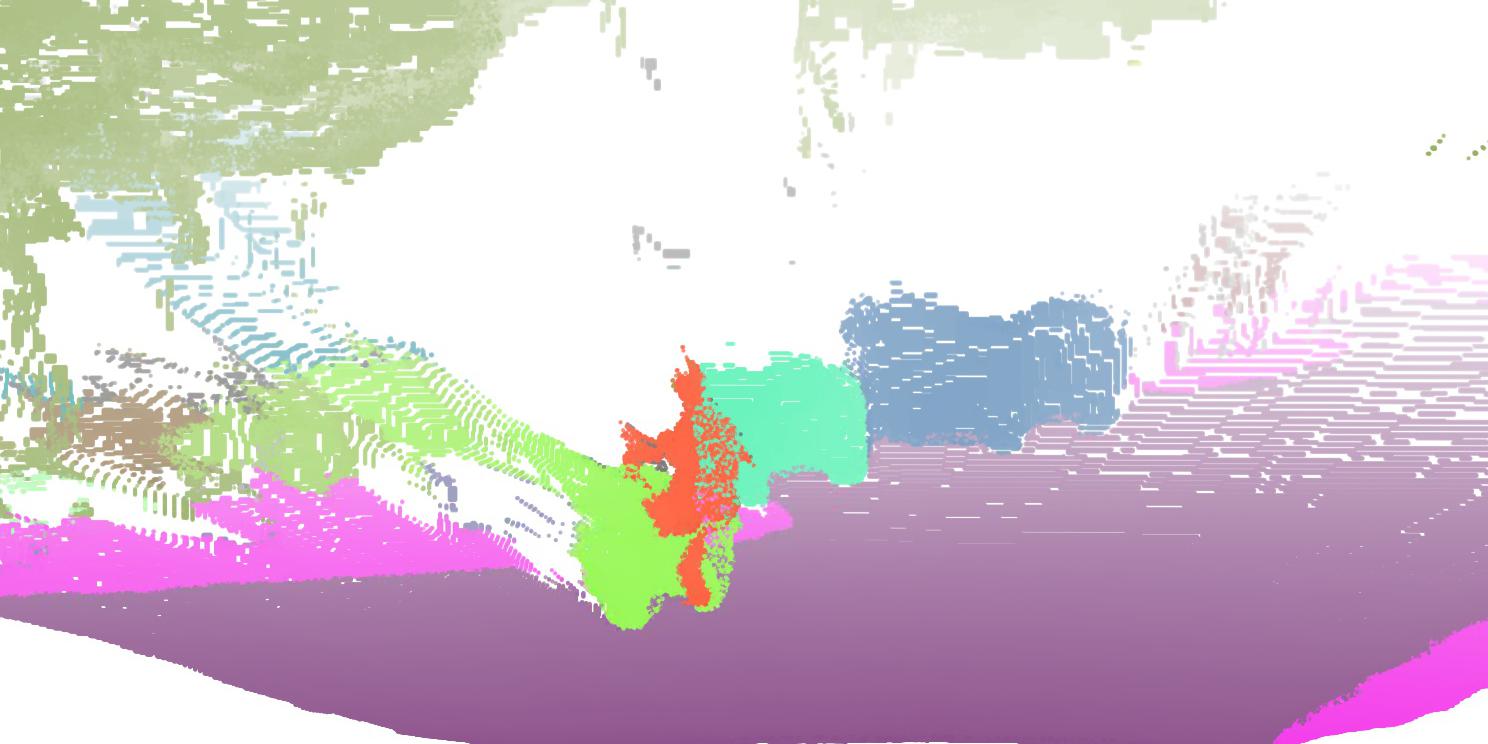}
\end{subfigure}\\\begin{subfigure}{.245\linewidth}
  \centering
  \includegraphics[trim={0 100 0 100},clip,width=\linewidth]{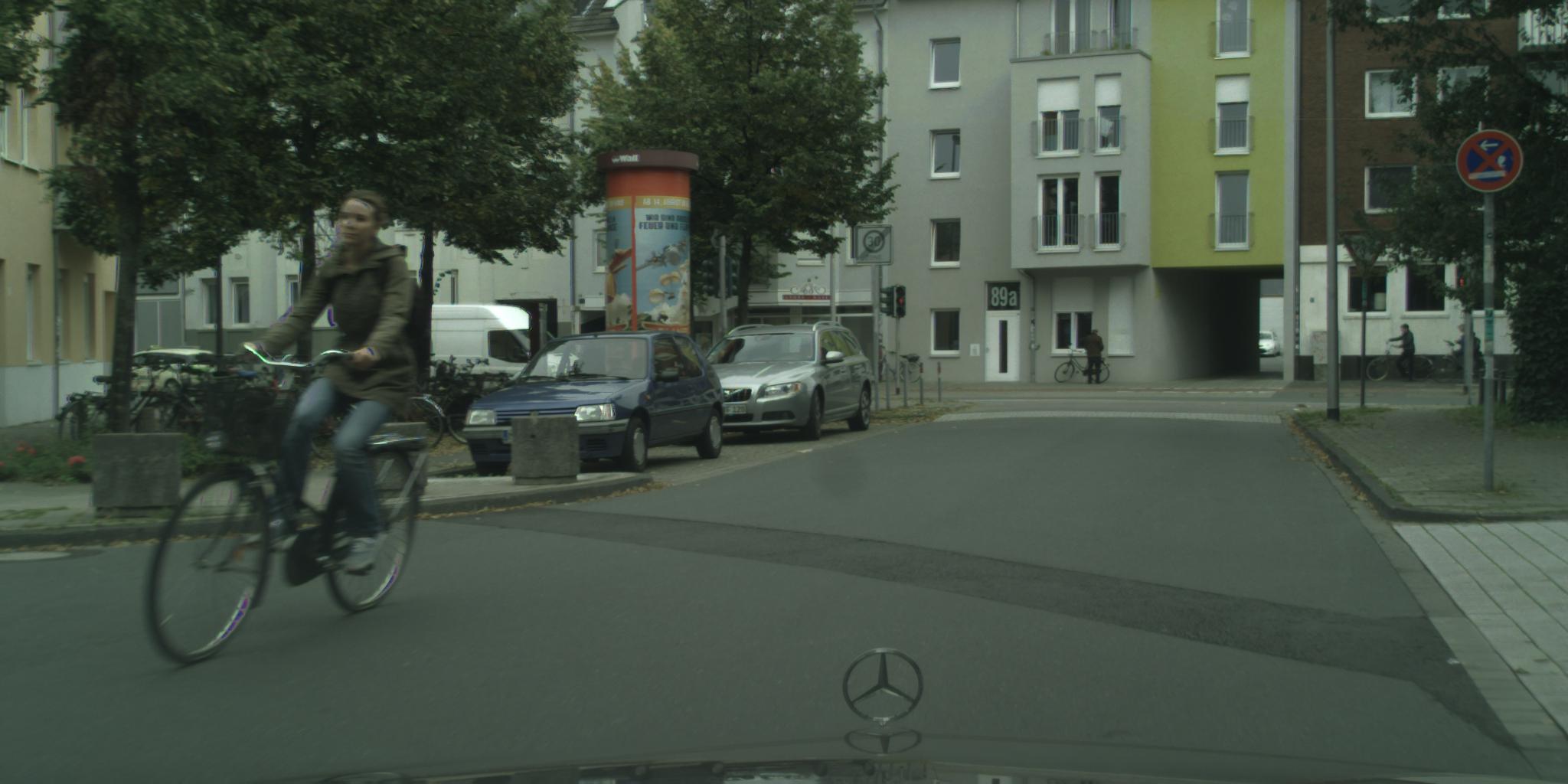}
\end{subfigure}\hfill\begin{subfigure}{.245\linewidth}
  \centering
  \includegraphics[trim={0 100 0 100},clip,width=\linewidth]{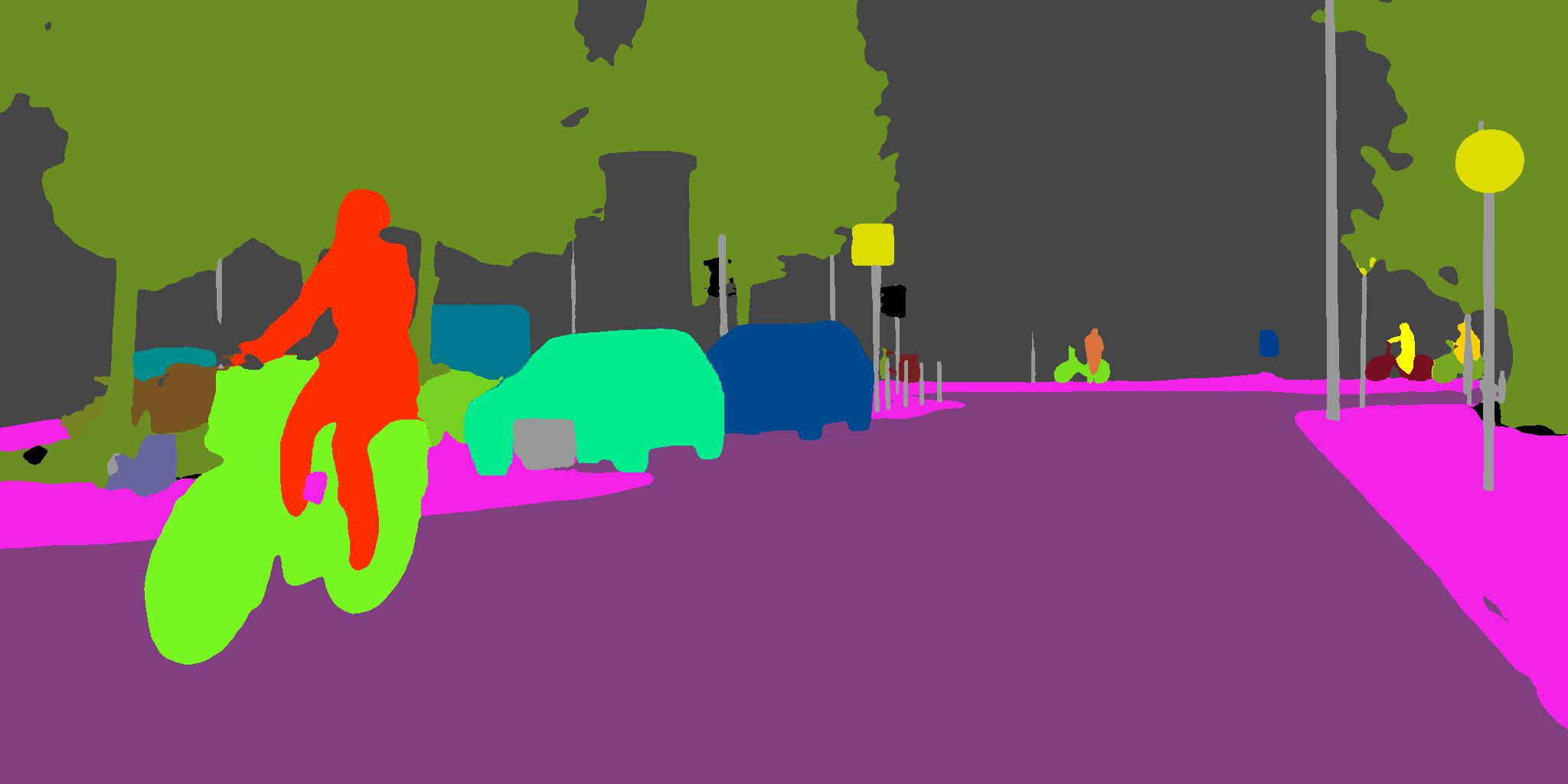}
\end{subfigure}\hfill\begin{subfigure}{.245\linewidth}
  \centering
  \includegraphics[trim={0 100 0 100},clip,width=\linewidth]{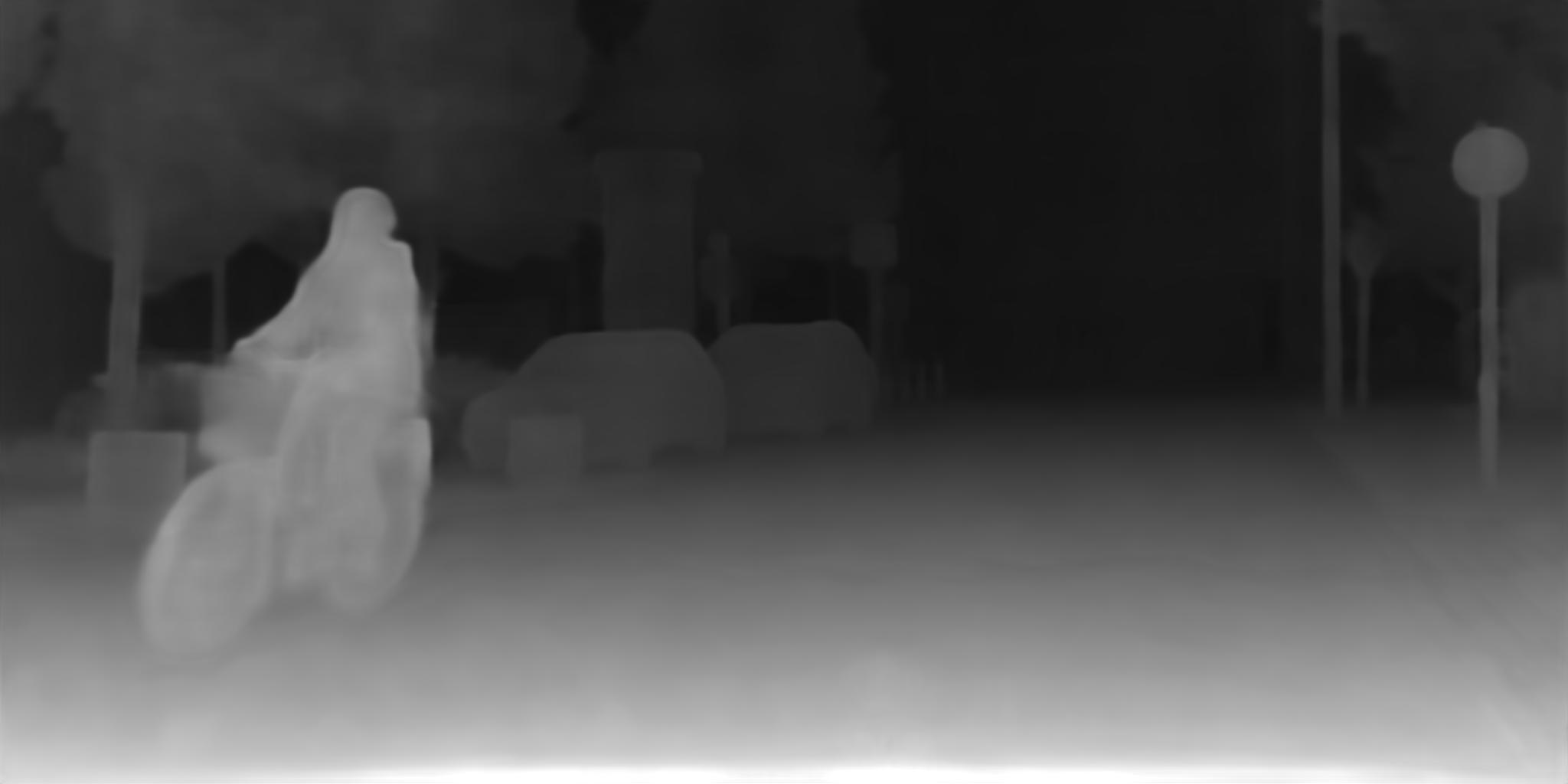}
\end{subfigure}\hfill\begin{subfigure}{.245\linewidth}
  \centering
  \includegraphics[trim={400 130 0 200},clip,width=\linewidth]{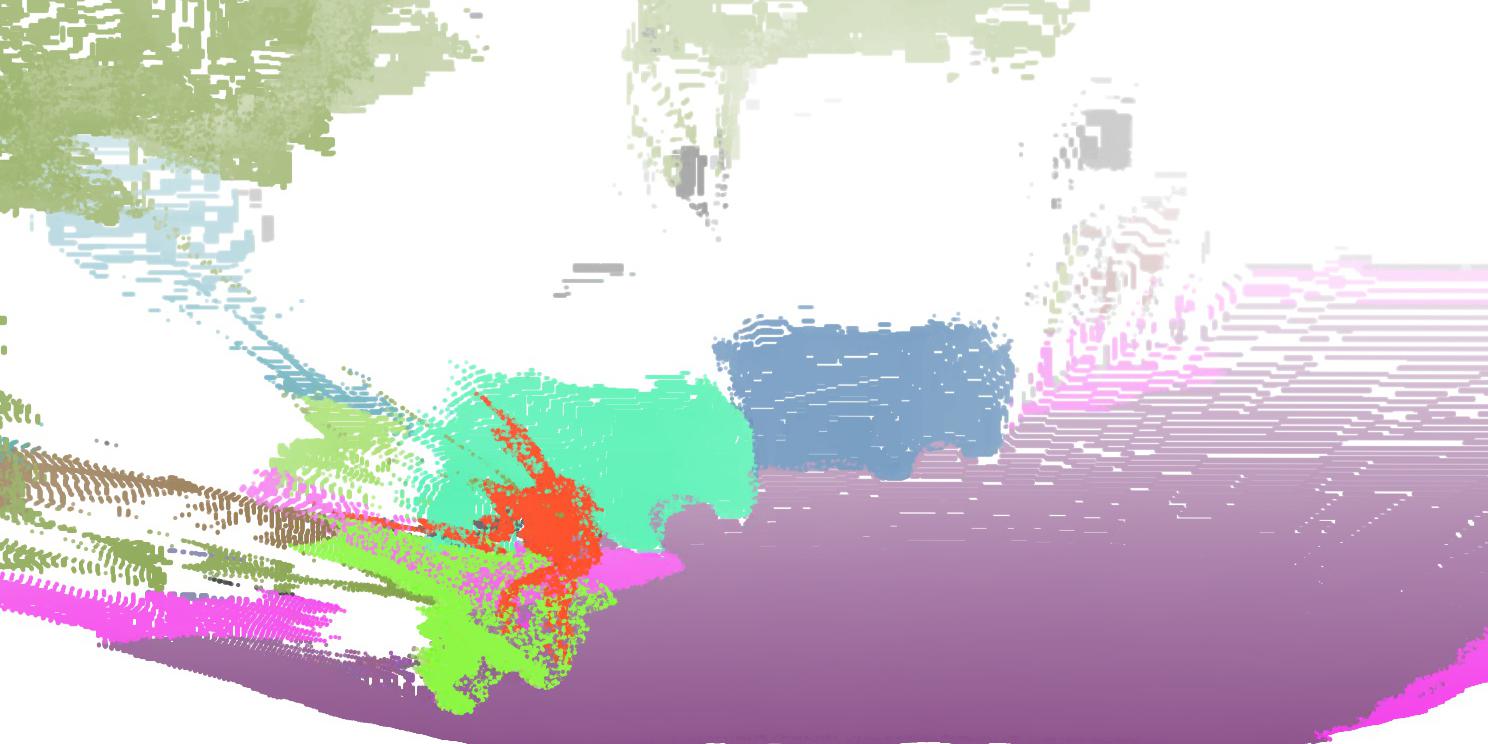}
\end{subfigure}\\
\caption{Prediction visualizations by ViP-DeepLab~\cite{qiao2021vip} on Cityscapes-DVPS~\cite{kim2020video}. From left to right: input image, temporally consistent panoptic segmentation prediction, monocular depth prediction, and point cloud visualization.}
\label{fig:cs_2}
\end{figure*}

\begin{figure*}[!th]
\centering
    \setlength\tabcolsep{1.12345pt}
    \begin{tabular}{cccccc}
        \includegraphics[width=0.156\textwidth,height=0.208\textwidth]{figs/image/12748.jpg} &
        \includegraphics[width=0.156\textwidth,height=0.208\textwidth]{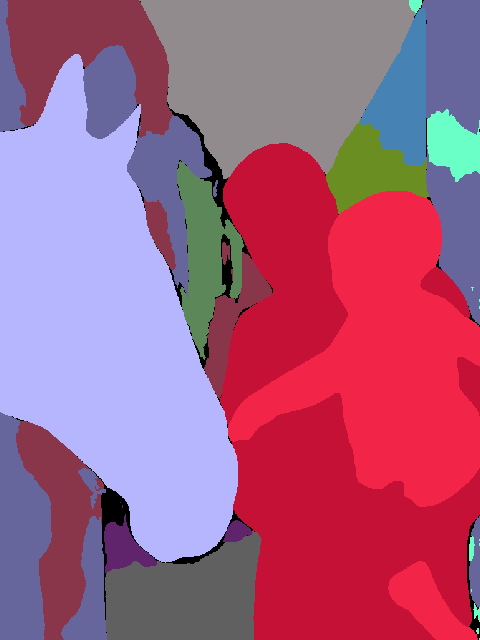} \hspace{0.5pt} & \hspace{0.5pt}
        \includegraphics[width=0.156\textwidth,height=0.208\textwidth]{figs/image/439525.jpg} &
        \includegraphics[width=0.156\textwidth,height=0.208\textwidth]{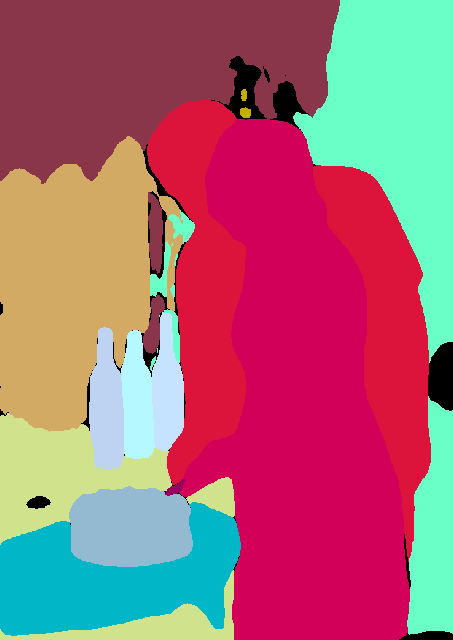} \hspace{0.5pt} & \hspace{0.5pt} \includegraphics[width=0.156\textwidth,height=0.208\textwidth]{figs/image/376322.jpg} &
        \includegraphics[width=0.156\textwidth,height=0.208\textwidth]{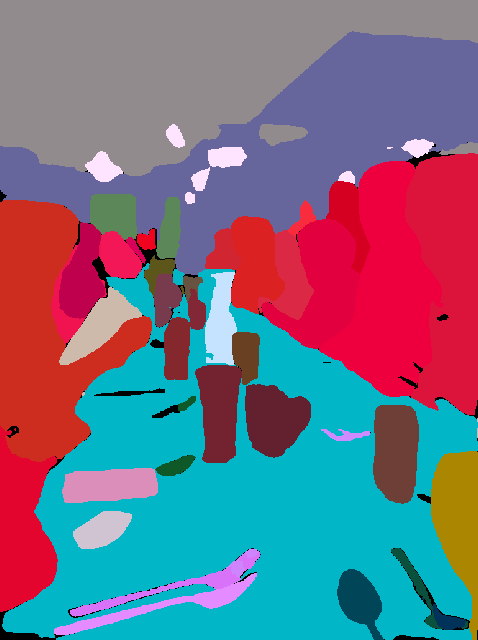} \\
        \includegraphics[width=0.156\textwidth,height=0.103\textwidth]{figs/image/127517.jpg} &
        \includegraphics[width=0.156\textwidth,height=0.103\textwidth]{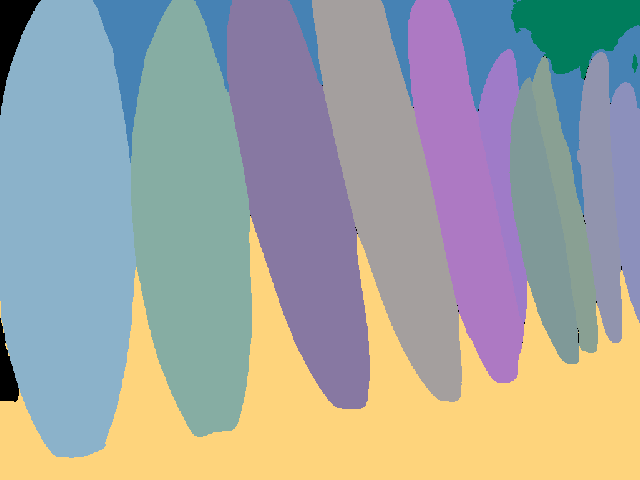} \hspace{0.5pt} & \hspace{0.5pt}
        \includegraphics[width=0.156\textwidth,height=0.103\textwidth]{figs/image/124975.jpg} &
        \includegraphics[width=0.156\textwidth,height=0.103\textwidth]{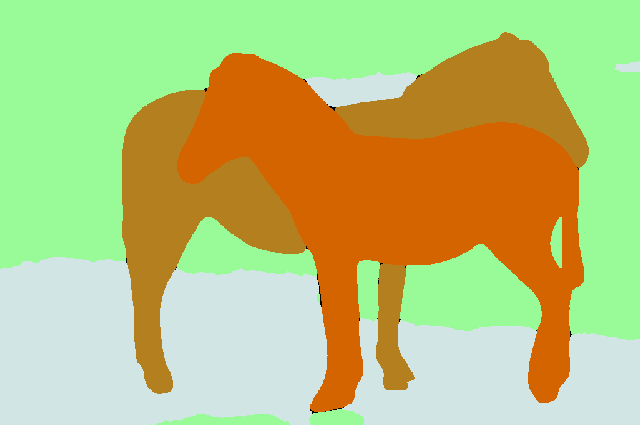} \hspace{0.5pt} & \hspace{0.5pt}
        \includegraphics[width=0.156\textwidth,height=0.103\textwidth]{figs/image/533493.jpg} &
        \includegraphics[width=0.156\textwidth,height=0.103\textwidth]{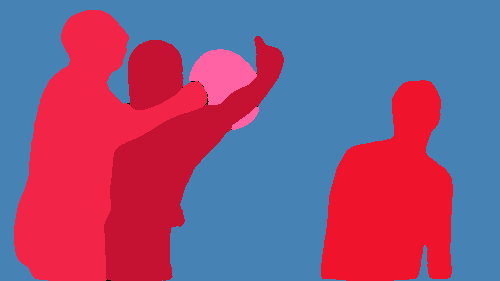} \\
        \includegraphics[width=0.156\textwidth,height=0.117\textwidth]{figs/image/109055.jpg} &
        \includegraphics[width=0.156\textwidth,height=0.117\textwidth]{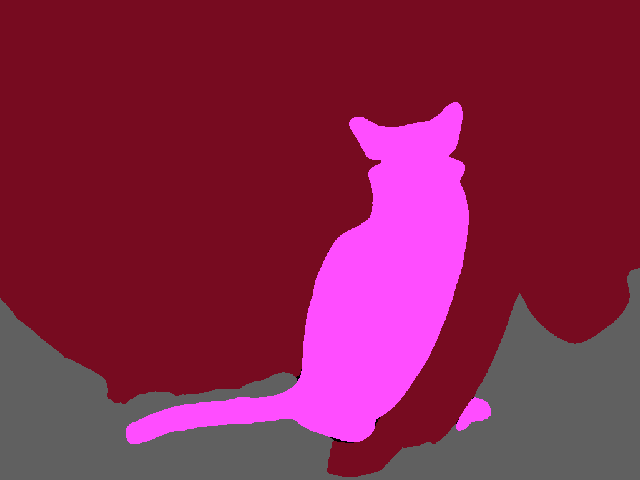} \hspace{0.5pt} & \hspace{0.5pt}
        \includegraphics[width=0.156\textwidth,height=0.117\textwidth]{figs/image/561958.jpg} &
        \includegraphics[width=0.156\textwidth,height=0.117\textwidth]{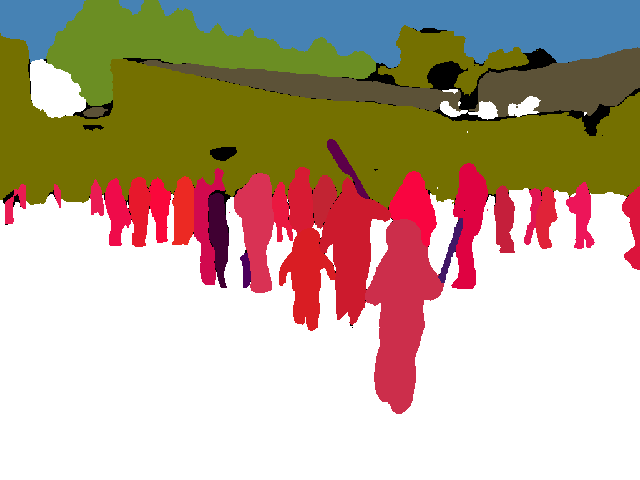} \hspace{0.5pt} & \hspace{0.5pt}
        \includegraphics[width=0.156\textwidth,height=0.117\textwidth]{figs/image/486104.jpg} &
        \includegraphics[width=0.156\textwidth,height=0.117\textwidth]{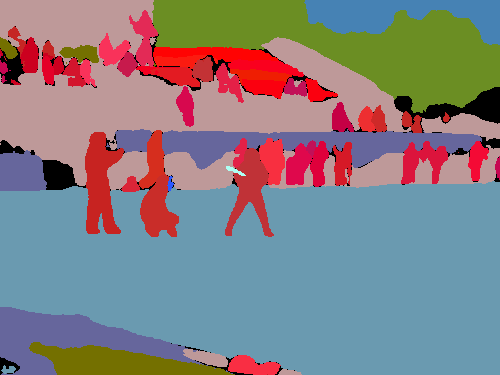}
    \end{tabular}%
    \caption{MaX-DeepLab~\cite{wang2021max} panoptic segmentation visualizations on the COCO~\cite{lin2014microsoft} {\it val} set.}
\label{fig:max_deeplab_results}
\end{figure*}

\begin{itemize}
\item \textbf{Image Semantic Segmentation}~\cite{he2004multiscale,ladicky2009associative,everingham2010pascal,mottaghi_cvpr14,chen2015attention,cordts2016cityscapes,zhou2019semantic}, one step further than image-level  classification~\cite{ILSVRC15} for scene understanding, recognizes objects within an image with pixel-level accuracy, requiring precise outline of objects. It is usually formulated as pixel-wise classification~\cite{long2014fully,deeplabv12015}, where each pixel is labeled by a predicted value encoding its semantic class.

\item \textbf{Image Instance Segmentation}~\cite{hariharan2014simultaneous,lin2014microsoft} recognizes and localizes object instances with pixel-level accuracy in an image. Existing models~\cite{dai2016instance,li2017fully,he2017mask,liu2018path,chen2019hybrid} are mostly based on the top-down approach (\ie, bounding box detection followed by segmentation) and formulate the problem as mask detection (one step further than bounding box detection for instance-level understanding). On the contrary, our system tackles instance segmentation from the bottom-up perspective, detecting (or more precisely, grouping) instances on top of segmentation prediction. Therefore, our system labels each `thing' pixel by a predicted value encoding both semantic class and instance identity (and the `stuff' pixels are ignored). Note that our model generates \textit{non-overlapping} instance masks, unlike other proposal-based models.

\item \textbf{Image Panoptic Segmentation}~\cite{kirillov2018panoptic} unifies semantic segmentation and instance segmentation. The task disallows \textit{overlapping} instance masks and requires labeling each pixel (including `thing' and `stuff' pixels) with a predicted value encoding both semantic class and instance identity. We would like to highlight that our whole system, including the training and evaluation pipelines, uses the panoptic label format (\ie, $panoptic\_label = semantic\_label \times label\_divisor + instance\_id$), and thus do not take (during training mode) or generate (during inference mode) any \textit{overlapping} masks~\cite{cheng2019panopticworkshop}. This is very different from most existing modern panoptic segmentation models~\cite{kirillov2019panoptic,xiong2019upsnet,li2018attention,porzi2019seamless,li2020unifying,mohan2020efficientps,Weber2020IROS,qiao2021detectors}, which are trained with \textit{overlapping} instance masks.

\item \textbf{Monocular Depth Estimation}~\cite{saxena2005learning,Silberman:ECCV12,geiger2013vision,eigen2014depth,fu2018deep} attempts to understand the 3D geometry of a scene by labeling each pixel with an estimated depth value.

\item \textbf{Video Panoptic Segmentation}~\cite{kim2020video,weber2021step} extends the image panoptic segmentation to the video domain, where a temporally consistent instance identity is enforced across the video sequence.

\item \textbf{Depth-aware Video Panoptic Segmentation}~\cite{behley2019semantickitti,qiao2021vip} provides in-depth scene understanding by solving a joint task of depth estimation, panoptic segmentation, and pixel-level tracking. Each pixel in a video is labeled with semantic class, temporally consistent instance identity, and estimated depth value.

\end{itemize}

%% file: sections/3.model.tex
\section{Model Garden}

Herein, we briefly introduce our developed DeepLab model variants that are included in the DeepLab2 library.

\begin{itemize}
\item \textbf{DeepLab}~\cite{deeplabv12015,chen2018deeplabv2,chen2017deeplabv3,deeplabv3plus2018}, where atrous convolution~\cite{deeplabv12015} (also known as convolution with holes or dilated convolution\footnote{\url{https://www.tensorflow.org/api_docs/python/tf/nn/atrous_conv2d}}) is intensively exploited for semantic segmentation (see several previous works ~\cite{holschneider1989real,giusti2013fast,papandreou2014untangling,sermanet2014overfeat} that effectively use atrous convolution in different ways). Specifically, DeepLabv1~\cite{deeplabv12015} employs atrous convolution to explicitly control the feature resolution computed by convolutional neural network backbones~\cite{sermanet2014overfeat,simonyan2014very,he2016deep,chollet2017xception,dai2017coco}. Additionally, atrous convolution allows us to effectively enlarge the model's field of view without increasing the number of parameters. As a result, the proposed ASPP module (Atrous Spatial Pyramid Pooling) in the follow-up DeepLab models~\cite{chen2018deeplabv2,chen2017deeplabv3,deeplabv3plus2018} effectively aggregates multi-scale information by employing parallel atrous conovlutions with multiple sampling rates.

\item \textbf{Panoptic-DeepLab}~\cite{cheng2019panoptic}, a simple, fast, and strong bottom-up (\ie, proposal-free) baseline for panoptic segmentation. Panoptic-DeepLab adopts the dual-ASPP and dual-decoder structures specific to semantic and instance segmentation, respectively. The semantic segmentation branch is the same as DeepLab, while the instance segmentation branch is class-agnostic, involving a simple instance center  regression~\cite{kendall2018multi,papandreou2018personlab,uhrig2018box2pix,neven2019instance,yang2019deeperlab}. Even though simple, Panoptic-DeepLab yields state-of-the-art performance on multiple panoptic segmentation benchmarks.

\item \textbf{Axial-DeepLab}~\cite{wang2020axial}, building on top of the proposed Axial-ResNet~\cite{wang2020axial} backbones that efficiently capture long-range context with precise position information. Axial-ResNet enables a huge or even global receptive field in all the layers of a backbone by replacing spatial convolutions with axial-attention layers~\cite{ho2019axial,huang2019ccnet} sequentially applied to the height- and width-axis. Additionally, a position-sensitive self-attention formulation is proposed to preserve context position in the huge receptive field. As a result, Axial-DeepLab, employing the proposed Axial-ResNet as backbone in the Panoptic-DeepLab framework ~\cite{cheng2019panoptic}, outperforms state-of-the-art convolutional counterparts on multiple panoptic segmentation benchmarks.

\item \textbf{MaX-DeepLab}~\cite{wang2021max}, the first fully {\it end-to-end} system for panoptic segmentation. MaX-DeepLab {\it directly} predicts a set of segmentation masks and their corresponding semantic classes with a mask transformer, removing the needs for previous hand-designed modules (\eg, box anchors~\cite{he2017mask}, thing-stuff merging heuristic~\cite{kirillov2019panoptic} or module~\cite{xiong2019upsnet}). The mask transformer is trained with a proposed PQ-style loss function and employs a \textit{dual-path} architecture that enables an Axial-ResNet to read and write a global memory, allowing efficient communication (feature information exchange) between any Axial-ResNet layer and the transformer. MaX-DeepLab achieves the state-of-the-art panoptic segmentation results on COCO~\cite{lin2014microsoft}, outperforming both proposal-based and proposal-free approaches.

\item \textbf{Motion-DeepLab}~\cite{weber2021step}, a unified model for the task of video panoptic segmentation, which requires to segment and track every pixel. It is built on top of Panoptic-DeepLab and uses an additional branch to regress each pixel to its center location in the previous frame. Instead of using a single RGB image as input, the network input contains two consecutive frames, \ie, the current and previous frame, as well as the center heatmap from the previous frame \cite{zhou2020tracking}. The output is used to assign consistent track IDs to all instances throughout a video sequence.

\item \textbf{ViP-DeepLab}~\cite{qiao2021vip}, a unified model that jointly tackles monocular depth estimation and video panoptic segmentation.
It extends Panoptic-DeepLab~\cite{cheng2019panoptic} by adding a depth prediction head to perform monocular depth estimation and a next-frame instance branch to generate panoptic predictions with temporally consistent instance IDs for videos.
ViP-DeepLab achieves state-of-the-art performance on multiple benchmarks, including KITTI monocular depth estimation~\cite{geiger2013vision,uhrig2017sparsity}, KITTI multi-object tracking and segmentation~\cite{geiger2013vision,voigtlaender2019mots}, and Cityscapes video panoptic segmentation~\cite{cordts2016cityscapes,kim2020video}.
\end{itemize}

%% file: sections/4.backbone.tex
\section{Supported Network Backbones}
In this section, we briefly introduce the network backbones that are supported by the DeepLab2 library.

\begin{itemize}
    \item \textbf{MobileNetv3}~\cite{howard2019searching}, a light-weight backbone designed for mobile devices~\cite{howard2017mobilenets,sandler2018mobilenetv2}, which thus could be used as a baseline for fast on-device model comparison.
    
    \item \textbf{ResNet}~\cite{he2016deep}, along with some modern modifications (\eg, Inception stem~\cite{szegedy2016rethinking}, stochastic drop path~\cite{huang2016deep}). This backbone could be used as a general baseline.

    \item \textbf{SWideRNet}~\cite{chen2020scaling}, which scales Wide ResNets~\cite{wrn2016wide,wu2019wider,chen2020naive} in both width (number of channels) and depth (number of layers). This could be used for strong model comparison.
    
    \item \textbf{Axial-ResNet}~\cite{wang2020axial} and \textbf{Axial-SWideRNet}~\cite{wang2020axial,chen2020scaling}, which replaces ResNet~\cite{he2016deep} and SWideRNet~\cite{chen2020scaling} residual blocks with axial-attention~\cite{ho2019axial,huang2019ccnet,wang2020axial} blocks in the last few stages. This hybrid CNN-Transformer (more precisely, only self-attention modules as the Transformer encoder) architecture could be used as a baseline for self-attention based model comparison. Besides the default axial-attention blocks, we also support global or non-local attention~\cite{vaswani2017attention,wang2018non} blocks.

    \item \textbf{MaX-DeepLab}~\cite{wang2021max}, which incorporates transformer blocks~\cite{vaswani2017attention} to Axial-ResNets~\cite{wang2020axial} in a dual-path manner, allowing efficient communication between any Axial-ResNet layers and the transformers. When turning off the transformer path, this backbone could be also adopted in the Panoptic-DeepLab framework~\cite{cheng2019panoptic}.

\end{itemize}

To briefly highlight the effectiveness of DeepLab2, our Panoptic-DeepLab employing the Axial-SWideRNet as network backbone achieves 68.0\% PQ or 83.5\% mIoU on Cityscapes validation set, with \textit{single-scale} inference and ImageNet-1K pretrained checkpoints. For more detailed results, we refer the readers to the provided model zoo\footnote{ \url{https://github.com/google-research/deeplab2/tree/main/g3doc/projects}}. 

Finally, we would like to emphasize that we have implemented the Axial-Block in a general way that encompasses not only transformer-based blocks (\ie, axial-attention~\cite{wang2020axial}, global-attention~\cite{vaswani2017attention}, dual-path transformer~\cite{wang2021max}) but also convolutional residual blocks (\ie, basic block, bottleneck block~\cite{he2016deep}, wide residual block~\cite{chen2020scaling}, each w/ or w/o Switchable Atrous Convolution~\cite{qiao2021detectors}). Our design allows users to easily develop novel neural networks that efficiently combine convolution ~\cite{lecun1998gradient}, attention~\cite{bahdanau2015neural}, and transformer~\cite{vaswani2017attention} (\ie, self- and cross- attention) operations.

%% file: sections/5.training.tex
\section{Data Augmentation During Training}
In addition to the typical data augmentation (\ie random scaling, left-right flipping, and random cropping) used for dense prediction tasks, we also support:

\begin{itemize}
\item A random color jittering found by AutoAugment~\cite{cubuk2018autoaugment}. In ~\cite{chen2020scaling}, we apply this augmentation policy with magnitude $1.0$, and $0.2$ on COCO and Cityscapes datasets, respectively.
\end{itemize}

%% file: sections/6.conclusion.tex
\section{Conclusion}
We open-source DeepLab2, containing all our recent research results, in the hope that it would facilitate future research on dense prediction tasks. The codebase is still under active development, and any contribution to the codebase from the community is very welcome. Finally, to reiterate, the code along with model zoo could be found at \url{https://github.com/google-research/deeplab2}.

\paragraph{Acknowledgments.}
We would like to thank Michalis Raptis for the feedback on the paper, Jiquan Ngiam and Amil Merchant for Hungarian Matching implementation, and the support from Google Mobile Vision.